\documentclass{article}
\usepackage{algorithm}
\usepackage{algpseudocode}
\usepackage{amssymb}
\usepackage{booktabs}
\usepackage{comment}
\usepackage{amsmath, bm}
\usepackage{tikz}
\usepackage{algorithmicx}
\usepackage{enumitem}
\usepackage{algpseudocode}
\usepackage{comment}
\usepackage{pgfplots}
\pgfplotsset{compat=1.17}  % or use a version that matches your setup
\usepackage{pgfplotstable}
\usepackage[toc,page]{appendix}
\usepackage{microtype,wrapfig}
\newcommand\blfootnote[1]{%
  \begingroup
  \renewcommand\thefootnote{}\footnote{#1}%
  \addtocounter{footnote}{-1}%
  \endgroup
}

\usepackage{subcaption} 
\usepackage[font=small,labelfont=bf]{caption}
\usepackage[final]{corl_2025} 
\newcommand{\algname}{\textsc{space2time}}
\title{From Space to Time: Enabling Adaptive Safety with Learned Value Functions via Disturbance Recasting}

\author{
\begin{tabular}[t]{@{\hspace{3em}}c@{\hspace{3em}}c@{\hspace{3em}}c}
  Sander Tonkens$^{\dagger}$ & 
  Nikhil Uday Shinde$^{\dagger}$ &
  Azra Begzadić$^{\dagger}$
\end{tabular}
\\
\begin{tabular}[t]{@{\hspace{3em}}c@{\hspace{3em}}c@{\hspace{3em}}c}
  \textbf{Michael C. Yip} &
  \textbf{Jorge Cortés} &
  \textbf{Sylvia L. Herbert}
\end{tabular}
\vspace{0.5em}
\\
\begin{tabular}[t]{c}
  \textnormal{University of California San Diego}
\end{tabular}
\vspace{0.5em}
\\
\href{https://stonkens.github.io/space2time}{https://stonkens.github.io/space2time}
\vspace{-3em}
}

\newtheorem{theorem}{Theorem}[section]

\newtheorem{definition}[theorem]{Definition}

\newtheorem{assumption}{Assumption}

\allowdisplaybreaks
\begin{document}
\maketitle

\begin{abstract}
The widespread deployment of autonomous systems in safety-critical environments such as urban air mobility hinges on ensuring reliable, performant, and safe operation under varying environmental conditions.
One such approach, value function-based safety filters, minimally modifies a nominal controller to ensure safety. 
Recent advances leverage offline learned value functions to scale these safety filters to high-dimensional systems. 
However, these methods assume detailed priors on all possible sources of model mismatch, in the form of disturbances in the environment -- information that is rarely available in real world settings. 
Even in well-mapped environments like urban canyons or industrial sites, drones encounter complex, spatially-varying disturbances arising from payload-drone interaction, turbulent airflow, and other environmental factors. 
We introduce \algname{}, which enables safe and adaptive deployment of offline-learned safety filters under unknown, spatially-varying disturbances. 
The key idea is to reparameterize spatial variations in disturbance as temporal variations, enabling the use of precomputed value functions during online operation.
We validate \algname{} on a quadcopter through extensive simulations and hardware experiments, demonstrating significant improvement over baselines.
\end{abstract}


%===============================================================================
\section{Introduction}
\blfootnote{$^\dagger$ Equal contribution. Correspondance to \href{stonkens@ucsd.edu}{stonkens@ucsd.edu}, \href{nshinde@ucsd.edu}{nshinde@ucsd.edu}}
Autonomous systems are increasingly deployed in safety-critical environments subject to variable conditions, where ensuring reliable and safe operation is of paramount importance. 
For instance, a drone operating in mapped environments such as urban canyons or shipyards must remain within a known safe region despite complex, spatially-varying wind disturbances. 
Rather than designing a bespoke performant, yet safe, controller for each task, a more modular approach uses a safety filter. 
These filters monitor a nominal, high-performance controller in real-time and intervene minimially-only when necessary to enforce guarantees without unduly compromising task performance~\cite{Hsu2023TheSF}. 
Popular approaches for constructing such filters include Control Barrier Functions (CBFs)~\cite{Ames2017ControlBF} and Hamilton-Jacobi Reachability (HJR) analysis~\cite{Bansal2017HamiltonJacobiRA}.
A recent line of work merges these two paradigms, leveraging reachability-based value functions as barrier certificates to construct safety filters with formal guarantees~\cite{ChoiLeeEtAl2021, TonkensHerbert2022, Begzadic2025BackTB}.

However, these methods face significant practical challenges. A primary limitation is their reliance on an accurate, pre-specified model of the system's dynamics and its operational domain-the set of conditions, such as expected wind patterns, the system is designed to operate in. 
Second, each method faces inherent hurdles: HJR analysis is limited by the curse of dimensionality, making it intractable for high-dimensional systems, while the systematic synthesis of a valid CBF for general nonlinear systems remains an open problem. 

To overcome these practical limitations, learning-based approaches have gained prominence, seeking to approximate safety value functions or barrier certificates directly from data~\cite{BansalTomlin2021, HsuRubies-RoyoEtAl2021, SoSerlinEtAl2024}. 
However, these learned approaches often assume a static operational domain that is known beforehand. 
This makes them brittle when faced with environmental conditions that shift during and across deployments, forcing a choice between unsafe behavior in the face of novelty or an overly conservative policy designed for the worst case~\cite{Nguyen2024GameplayFR}.

Offline learning of a value function for a safety filter relies on a joint system-environment model that captures the true system's runtime behavior. 
Such a model is infeasible in environments with spatially varying disturbances, e.g., wind in urban canyons~\cite{HUNTER1990315, AchermannLawranceEtAl2019}, which are unknown a priori and even differ across deployments. 
A compounding challenge arises because disturbance measurements are typically obtained at a slower rate than control inputs, due to practical sensing and computational constraints.
However, this slower update rate means unmodeled spatial variation can cause significant changes between consecutive measurements, leading to safety violations if ignored. 
Our insight is that spatial variations in disturbance appear as temporal variations along a trajectory. 
By learning a time-varying safety value function that explicitly accounts for disturbance evolution over time, we implicitly capture spatial variations along trajectories, enabling their use as online safety filters.
This work takes a step towards bridging offline-learned value functions with online adaptation in evolving operational domains. 
Our main contributions are:

\begin{figure}[t!] 
 \resizebox{\linewidth}{!}{\input{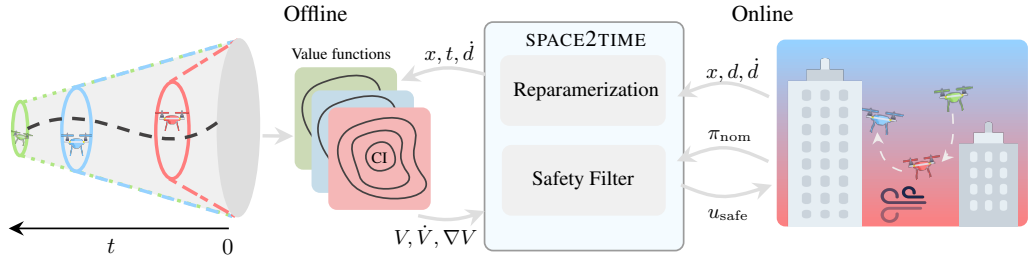}}
 \caption{Conceptual overview of the \algname{} framework. 
 \textbf{Offline}, we learn a family of safety value functions, each corresponding to a system evolving under a different time-varying disturbance profile. \textbf{Online}, we reparameterize an estimate of the disturbance and its derivative into its temporal equivalent to query the adaptive value function-based safety filter. 
 \algname{} ensures the system is \textit{realistic} about the present (based on the current disturbance) and \textit{pessimistic} about the future (assuming the worst-case rate of change to persist).
 }
 \label{fig:overview}
 \vspace{-0.5cm}
\end{figure}

\begingroup
\setlength{\itemsep}{0pt}
\setlength{\parskip}{0pt}
\begin{itemize}[leftmargin=*]
    \item We introduce a safety value function formulation that is explicitly conditioned on a disturbance's temporal rate of change.
    \item We use this value function formulation to propose \algname{}.
    Our approach reparameterizes spatial variations as temporal variations in disturbance. 
    This ensures safety in the presence of unknown, spatially varying disturbances through the use of an adaptive safety filter that leverages our offline-learned value functions. 
    \item We validate \algname{} on a quadcopter through extensive simulations and hardware experiments demonstrating substantial improvements in safety compared to existing approaches without significantly sacrificing performance. 
\end{itemize}
\endgroup

\section{Related Work}
\label{sec:citations}
The past decade has seen major progress in safety filtering for robotic systems, particularly through approaches grounded in HJR and CBFs. We highlight relevant approaches to our setting and point readers to recent surveys that explore safety filters and their trade-offs~\cite{brunke2022safe, Wabersich2023DataDrivenSF}.

\textbf{Learning-Based Approaches for Hamilton Jacobi Reachability analysis. }HJR produces a value function whose zero level set encodes the set of initial conditions from which a system may reach a goal and/or avoid an obstacle despite worst-case disturbance~\cite{Bansal2017HamiltonJacobiRA}. 
Traditional approaches to solve HJR rely on dynamic programming and therefore scale exponentially with the dimensionality of the system. 
In recent years, learning-based approaches have vastly improved the scalability of HJR.
Reinforcement learning-based methods have been developed to estimate HJR value function using a specific safety Bellman Equation~\cite{FisacLugovoyEtAl2019}, with success in many applications~\cite{HsuRubies-RoyoEtAl2021, Nguyen2024GameplayFR}. 
Physics-informed neural networks (PINNs) have also been employed in a self-supervised manner to approximate the value function~\cite{BansalTomlin2021}; an extension of this work learns a parameterized version of the reachability value function based on different disturbance bounds~\cite{Borquez2022ParameterConditionedRS}. 
This approach inspires our work, as it can provide safety across a range of operational domains, each with a different maximum disturbance level. 
However, the prior work assumes that the disturbance bound is constant over space and time; in an environment with spatially varying disturbances, this approach would not preserve safety.

\textbf{Learning-Based Approaches for Control Barrier Functions.} Safety is also frequently ensured through the use of Control Barrier Functions (CBFs)~\citep{Ames2017ControlBF}. 
When a CBF is available, synthesizing a safe controller involves formulating an optimization-based safety filter that minimally adjusts potentially unsafe control inputs and has been widely deployed in robotics~\cite{Tayal2023ControlBF,Yu2024EfficientMP,Grandia2020MultiLayeredSF}. 
The construction of a valid CBF and feasibility of the safety filter remains a challenge~\cite{brunke2022safe}. 
Recently, learning-based approaches have been introduced to obtain CBFs, but they often lack formal guarantees or rely on restrictive assumptions~\cite{Robey2020LearningCB,SoSerlinEtAl2024,Tayal2024LearningAF, pmlr-v270-manda25a}. 
As an alternative, recent work has shown that CBFs can be constructed directly using techniques from HJR \cite{ChoiLeeEtAl2021}. 

\section{Background} \label{sec:background}
Consider a control- and disturbance-affine system of the form:
\begin{equation}\label{eq:orig_sys}
    \dot{x} = \tilde{f}(x,u,d)=  f(x) + g(x)u +d,
\end{equation}
where $x \in \mathbb{R}^n$ is the state and $u \in \mathcal{U} \subseteq \mathbb{R}^p$ is the control input. The state-dependent disturbance $d \in \mathcal{D} \subseteq \mathbb{R}^q$ is determined by a disturbance field $w:\mathcal{X}\to\mathcal{D}$, such that at any state $x$, the disturbance value is $d=w(x)$. 
$\mathcal{U}$ and $\mathcal{D}$ are convex and compact sets. We consider a time horizon $[t,0]$, where the initial time $t\leq 0$. 
For each initial time $t \le 0$, we denote the sets of admissible control and disturbance signals by $\mathbb{U}(t) := \{\bm{u} : [t, 0] \to \mathcal{U} \mid \bm{u} \text{ is measurable}\}$ and $\mathbb{D}(t):=\{\bm{d} : [t, 0] \to \mathcal{D} \mid \bm{d} \text{ is measurable}\}$.

Throughout this work, we make the following assumption about the dynamics $\tilde{f}$.
\begin{assumption} \label{as:flowfield}
    The function $\tilde{f}: \mathbb{R}^n \times \mathcal{U} \times \mathcal{D} \to \mathbb{R}^n$ is above bounded by $M_{\tilde{f}}$ and globally Lipschitz.
\end{assumption}

Under this assumption, given an initial time and state $t$ and $x$, a control signal $\bm{u}$ and disturbance signal $\bm{d}$ there exists a unique solution $\bm{x}_{x,t}^{\bm{u},\bm{d}}:[t,0] \to \mathbb{R}^n$ of the system with initial condition $\bm{x}(t) = x$.

\textbf{Hamilton-Jacobi Reachability} (HJR) is a model-based optimal control framework that characterizes the set of initial states from which a system can reach a target set and/or avoid a failure set~\cite{Bansal2017HamiltonJacobiRA}.
Here, we consider both the avoid problem and the reach-avoid problem. For both problems, we consider a constraint function $g:\mathbb{R}^n\to\mathbb{R}$ describing the failure set $\mathcal{F} := \{ x \in \mathbb{R}^n \mid g(x)  \leq 0 \}$. Then, the reward function associated with the avoid problem is:
\begin{align}
    r_\text{A}(x,t,\bm{u}, \bm{d}) = \min_{\tau \in [t,0]}g(\bm{x}_{x,t}^{\bm{u}, \bm{d}}(\tau)).
\end{align}
This encodes the minimum value of  $g$ attained by the trajectory over the time horizon.  If this minimum is smaller than 0, the trajectory entered the failure set. For the reach-avoid problem, additionally consider a target function $l:\mathbb{R}^n\to\mathbb{R}$ describing the target set $\mathcal{T} := \{ x \in \mathbb{R}^n \mid l(x) \geq 0 \}$. 

Then, the reward function associated with the reach-avoid problem is:
\begin{align}
    r_{\text{RA}}(x,t,\bm{u}, \bm{d}) = \max_{\tau \in [t,0]}\min\{l(\bm{x}_{x,t}^{\bm{u},\bm{d}}(\tau)), \min_{s \in [t,\tau]}g(\bm{x}_{x,t}^{\bm{u}, \bm{d}}(s))\}.
\end{align}\label{eq:reward_ra}
A trajectory has a positive reward if it reaches the target $\mathcal{T}$ at some time $\tau_1$, i.e., $l(\bm{x}_{x,t}^{\bm{u},\bm{d}}(\tau_1))\geq0$ while having stayed clear of the failure set $\mathcal{F}$ until that time $\tau_1$, i.e., $g(\bm{x}_{x,t}^{\bm{u}, \bm{d}}(s))\geq  0$ for all $s \in [t, \tau_1]$. 

HJR considers a game in which one player, here the control $u$, tries to maximize the reward, whereas a second player, here the disturbance $d$, acts antagonistically and attempts to minimize the reward. The value function of this game is defined as $V(x,t)=\min_{d \in \mathcal{D}}\max_{u\in\mathcal{U}}r(x,t,\bm{u},\bm{d})$.

For the avoid problem, the value function encodes the avoid tube $\mathcal{A}(\mathcal{F},t):=\{x \in \mathbb{R}^n \mid V(x,t)\geq 0\}$, which is the set of states that can avoid the failure set for time $t$.

Alternatively, the reach-avoid tube $\mathcal{RA}(\mathcal{T}, \mathcal{F}, t):= \{ x \in \mathbb{R}^n \mid V(x,t) \geq 0 \}$ represents the set of states from which the system is guaranteed to safely reach the target while avoiding the failure set within time $t$. 
Appendix~\ref{sec:appendix_th_background} and~\ref{sec:appendix_learning_ra} discuss how the reach and reach-avoid problems can be solved directly with dynamic programming programming~\cite{Bansal2017HamiltonJacobiRA} and approximately with self-supervised learning~\cite{BansalTomlin2021}.

\textbf{Control Barrier Functions} (CBFs) provide a framework to prevent a system from entering the failure set. Typically, CBFs are used for control-affine systems without disturbances, i.e., $\dot{x}~=~f(x)~+~g(x)u$~\cite{Ames2017ControlBF}.
A continuously differentiable function $h$ is a CBF if 1) we can represent a given safe set $\mathcal{C}$ as the 0-superlevel set of $h$, and 2) there exists an extended class $\mathcal{K}$ function $\alpha$ such that, for each $x \in \mathcal{C}$, there exists $u \in \mathbb{R}^m$ satisfying $ \nabla h(x)^{\top} (f(x)+g(x)u)+\alpha(h(x)) \geq 0$. For details on this formulation, see Appendix~\ref{sec:appendix_th_background}.

If a Lipschitz continuous controller $k: \mathbb{R}^n \to \mathbb{R}^m$, defined as $u = k(x)$, satisfies the CBF constraint for all $x \in \mathcal{C}$, it can be used to ensure safety. 
Given any nominal control law $u_\text{nom}: \mathbb{R}^n \to \mathbb{R}^p$ that may violate safety, CBFs allow for minimal correction using the following optimization problem:
\begin{equation}\label{eq:qp}
u^*(x)=\arg \min _{u }\left\|u - u_{\text {nom }}(x)\right\|_2^2 \text { subject to } \nabla_{x} h(x)^\top (f(x)+g(x)u)+\alpha(h(x)) \geq 0.
\end{equation}
As we consider control-affine dynamics, such optimization problems are quadratic programs, resulting in a computationally efficient safety filter. 
The main challenge with CBFs is in finding a valid CBF. 
To address this, CBFs can be constructed using HJR~\cite{ChoiLeeEtAl2021, Begzadic2025BackTB}, which incorporates finite time horizons, control bounds, and disturbance bounds. 

\section{Problem Statement}
We consider safety‐filter synthesis for control- and disturbance-affine systems operating under spatially varying disturbances. This disturbance field $w(x):\mathcal X\to\mathcal{D}_{\max}$ is unknown, but its magnitude is bounded by $d_{\max} \in \mathbb R^q$, i.e. $\mathcal{D}_{\max}=\{d\in\mathbb{R}^q \mid \lvert d \rvert \leq d_{\max} \}$ and it is Lipschitz continuous with a 
known constant $L_d$. Given a prescribed failure set $\mathcal F\subset\mathcal X$, our objective is to design a safety filter that adapts to the spatially-varying disturbance, guarantees avoidance of $\mathcal F$, and admits an offline‐learnable value function for formal safety certification. We consider the following approaches that have been pursued in the literature for this problem formulation.

\textbf{Constant Maximum Disturbance (Worst Case).} Classical reachability-based formulations typically assumes fixed control and disturbance sets $\mathcal{U}$ and  $\mathcal{D}$~\cite{Bansal2017HamiltonJacobiRA}. 
While control inputs are often (in practice) spatially invariant across the environment, disturbances such as wind may vary more smoothly over space. 
Based on the operational design domain, a maximum allowable disturbance set $\mathcal{D}_{\max}$ is typically considered to account for worst-case conditions. 
However, naively assuming that the disturbance can take any value within $\mathcal{D}_{\max}$ at all states might result in a very conservative value function $V(x,t, \mathcal{U}, \mathcal{D}_\text{max})$, particularly if this maximum disturbance is localized in the state space. 

\textbf{Perfect Information (Oracle).} An oracle value function would have perfect access to the complete deterministic state-dependent disturbance field $w(x)\in\mathcal{D}_{\max}$, and compute the optimal value function with $\mathcal{D}_\text{oracle}(x)=\{w(x)\}$, to find $V_\text{oracle}=V(x,t,\mathcal{D}_\text{oracle})$. 
In practice, however, the disturbance field $w(x)$ is unknown prior to deployment making it unsuitable for offline learning. 
Pretraining for every possible spatially-varying disturbance landscape is intractable. 

\textbf{Parameterizing the Pretrained Value Function by Constant Disturbance Sets (Naive).} An alternative is to pretrain a value function for a range of potential maximum disturbances that are constant across space, i.e., an ensemble of $K$ value functions, where each member $k$ corresponds to a constant disturbance bound set $\mathcal{D}_k$ satisfying $\mathcal{D}_1 \subseteq \cdots \subseteq \mathcal{D}_K \subseteq \mathcal{D}_{\max}$. This yields a family of value functions $V_1(x,t), \dots, V_K(x,t)$ that can be used to approximate the true disturbance-dependent behavior~\cite{Borquez2022ParameterConditionedRS, Jeong2024ParameterizedFA}. 
It can be used to naively switch between the $K$ value functions based on the locally observed disturbance online. 
In place of an ensemble, one could also learn a single-value function directly parameterized by the disturbance bound. 
However, each value function assumes that the current disturbance bound will remain constant over space and time. 
In spatially-varying environments it therefore does not guarantee safety, which can induce safety-critical failures. 

\textbf{Fine-tuning the Value Function Online.} In contrast, for low-dimensional systems,~\cite{Fisac2017AGS, Herbert2021ScalableLO} propose iteratively collecting data to estimate (using Gaussian Processes~\cite{Rasmussen2006}) and update  the spatially varying disturbance set $\mathcal{D}(x)$ and compute the corresponding value function with HJR using dynamic programming (DP). 
It is therefore limited to low-dimensional systems, where DP is tractable. 
This approach, however, is not suitable for high-dimensional systems for which value functions are typically learned, as training (or finetuning) neural networks cannot occur in the loop.

Given the limitations of each approach described above, there is a need to formulate a value function that is amenable to offline learning, but can be used in an adaptive manner online in the face of changing disturbances while maintaining safety. This is the problem tackled in this paper.

\section{From Space to Time: Time-Varying Disturbances for Adaptive Safety}
\begin{wrapfigure}{r}{0.45\linewidth}
    \centering
    \vspace{-0.9cm}
\input{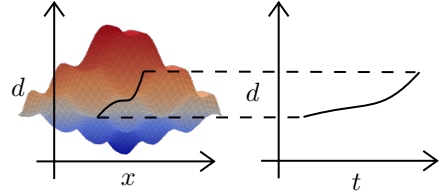}
    \caption{
    Change in disturbance bounds over state $x$ is encoded as change in bounds over time $t$.
    The disturbance bounds increase towards the red region in the left image. 
    This is encoded as a temporal disturbance increase shown in the plot on the right. 
    }
    \label{fig:method_time_to_space}
    \vspace{-0.3cm}
\end{wrapfigure}

\subsection{Offline Safety Value Function Learning}

Our key insight is that changes in disturbance bounds over space can be encoded as changes in disturbance bounds over time as the system moves through the environment as shown in Fig.~\ref{fig:method_time_to_space}. 
Typically, the operational domain is specified by both a maximum disturbance magnitude and a bound on the disturbance's maximum rate of change over space, specified by the disturbance Lipschitz constant $L_{d}$. 
For offline learning, we reformulate this spatial variation as a disturbance that grows linearly with time, along the system's trajectory. 
Specifically, the disturbance grows at a rate bounded by $\dot{d}_{\mathrm{max}}= L_d M_{\tilde{f}}$, where $M_{\tilde{f}}$ bounds the system dynamics $\tilde{f}$. 
By assuming these time-varying dynamics, where disturbances increase over time, we obtain robustness to spatially varying disturbances encountered along the trajectory. 
For notational simplicity, we first consider a one-dimensional disturbance, i.e., $d:\mathcal X\to\mathbb R$. 
Given a rate of change of a disturbance $\dot{d}$, we define the associated time-varying disturbance set as
\begin{equation}
    \mathcal{D}_\text{tv}(\dot{d},t)=\big\{d\in \mathbb{R} \mid \lvert d \rvert \leq \max\{0, d_{\max} - \lvert t \rvert \dot{d}\}\big\}.
\end{equation}
In practice, to reduce conservativeness and improve flexibility, we construct an ensemble of $K$ value functions, each associated with a constant disturbance rate $ \dot{d}_1 \leq \cdots \leq\dot{d}_K \leq \dot{d}_{\max}$ (or a value function parameterized by the disturbance rate). 
Intuitively, as the system moves forward in time, the disturbance set $\mathcal{D}_{\mathrm{tv}}$ gets larger. 
To capture this evolution, we augment the dynamics to explicitly model the disturbance rate, resulting in 
\begin{equation} \label{eq:sys_new_dotd}
\dot{z} = \hat{f}(z,u, \eta) =
\begin{bmatrix}
\dot{x} \\ \ddot{d}
\end{bmatrix}
=
\begin{bmatrix}
f(x)+g(x)u+ \eta\\ 0
\end{bmatrix},
\end{equation}
where $z=[x, \dot{d}]\in \mathbb{R}^n \times \mathcal{\dot{D}}$, with $\mathcal{\dot{D}}=\{\dot{d} \in \mathbb{R}^q \mid 0 \leq \dot{d}\leq \dot{d}_\text{max}\} $, is the joint state and $\eta\in \mathcal{D}_{\text{tv}}(\dot{d}, t)$ is the disturbance subject to time-varying bounds. 
This formulation is particularly relevant for robotic systems operating in environments where disturbances vary smoothly over space as the robot navigates through them (thus also smoothly over time), such as aerial vehicles encountering structured wind fields, or ground robots traversing regions with gradual changes in terrain or material properties. 

Our proposed approach requires introducing a target set $\mathcal{T}$ which is control-invariant under worst-case disturbance, i.e., for all $x \in \mathcal{T}$ there exists a control input $u \in \mathcal{U}$ such that the system remains in $\mathcal{T}$ for all disturbances $d\in\mathcal{D}_\text{max}$.  
Such a set can be learned a priori or expertly-chosen, see e.g.,~\cite{Nguyen2024GameplayFR}. 
This set can be interpreted as a fallback region, such as a docking location or an open-sky area above a city. 
Having designed a constraint function $g$ describing the environment's failure set $\mathcal{F}$ and a target function $l$ for the above-defined target set $\mathcal{T}$, we can compute or learn the value function $V(z,t)$ associated with the reach-avoid game~\eqref{eq:reward_ra} under dynamics~\eqref{eq:sys_new_dotd}. 
In the one-dimensional case, the disturbance magnitude $d$ is determined by the time $t$ and disturbance rate $\dot{d}$ through the relation $d=d_{\max} - \lvert t \rvert\dot{d}$. Thus, for the resulting value function, its arguments implicitly capture the current disturbance magnitude. 

The disturbance-rate parameterized value function $V(z,t)$ for the dynamics in~\eqref{eq:sys_new_dotd} can be effectively learned offline using the self-supervised methods outlined in Appendix~\ref{sec:appendix_learning_ra}. 
The design of a safety filter based on the proposed value function and its online application is presented in the subsequent section.

\subsection{\algname{}: Online Deployment of Temporally-Varying Value Functions}

\begin{wrapfigure}{r}{0.5\textwidth}
    \vspace{-24pt}
    \begin{minipage}{0.5\textwidth}
        \begin{algorithm}[H]
            \caption{\algname{}}\label{alg:cap}
            \begin{algorithmic}[1]
                \Require $V(z,t), \> \pi_\text{nom}(x)$
                \State Measure state $x$
                \State Update $\bar{d}, \overline{D_{\tilde f}d}$ (Measured at a slower rate)
                \State $t_\text{return}\gets \min\left\{\frac{d_{\max} - \bar{d}}{\overline{D_{\tilde f} d}}\right\}$,
                $z\gets [x,\overline{D_{\tilde f} d}]$
                \State $\dot{V}, \nabla_z V, V \gets \text{NN}(V(z,t_\text{return})\textbf{)}$
                \State $u^* \gets \textbf{CBF(}z,V,\frac{\partial}{\partial t} V, \nabla_z V, \pi_{\text{nom}}(x)\textbf{)}$
                \State Apply $u^*$ to system.
            \end{algorithmic}
        \end{algorithm}
    \end{minipage}
    \vspace{-14pt}
\end{wrapfigure}

Our objective is designing a safety filter based on the ensemble of reach-avoid value functions for dynamics~\eqref{eq:sys_new_dotd} to ensure safety online for the system subject to the dynamics~\eqref{eq:orig_sys} with spatially varying disturbances.  
We consider a setting where disturbances $d$ and its directional derivative with respect to the dynamics $D_{\tilde f}{d}={\nabla_x d}^\top \tilde f$ can be estimated 
online\footnote{In practice, the directional derivative is approximated through a finite difference approach, i.e. $D_{\tilde f}{d} \approx \frac{\Delta d}{\Delta t}.$}. 
Recall $\dot{d}_\text{max}=L_d M_{\tilde f}$, where $L_d$ is the Lipschitz constant associated with the disturbance field $w(x)$ and $M_{\tilde f}$ upper-bounds the dynamics. 
Then, $\lvert D_{\tilde f}{d}\rvert\leq \dot{d}_\text{max}$. 
An overly cautious, yet verifiably safe, safety filter could then consider the worst-case scenario at all times by querying the value function at the maximum disturbance rate $\dot{d}_\text{max}$ at all states $x$ and its associated estimated disturbance $d$. 
While this ensures that safety is preserved regardless of the actual disturbance evolution, it is very conservative and impedes the system's nominal objectives.

Instead, we propose an adaptive strategy that leverages estimates of both the disturbance $d$ and its directional flow $D_{\tilde f}{d}$. 
We store the past $H$ estimated values for $D_{\tilde f}{d}$, and select the maximum (per dimension) value over the horizon $H$ as our  directional derivative sample, while using the most recent estimate of $d$. 
Given estimates $\bar{d}$ and $\overline{D_{\tilde{f}}d}$, the time to return to the control invariant set $\mathcal{T}$ is given by: 
\begin{equation}
    t_{\text{return}} = \min\left( \frac{d_{\max} - \bar{d}}{\overline{D_{\tilde{f}}d}} \right),
\end{equation}
and the value at the current state is given by $V(z, t_\text{return})$, with $z=[x, \overline{D_{\tilde{f}}d}]$
Then, given a nominal controller $u_\text{nom}=\pi_\text{nom}(x)$, we synthesize a control input $u^*$ using the VB-CBF-based safety filter~\cite{Begzadic2025BackTB}:
\begin{equation}\label{eq:qp_vb_cbf}
\begin{aligned}
u^*(z, t_\text{return}) =&\arg \min _{u \in \mathcal{U}}\left\|u - u_\text{nom}\right\|_2^2 \\
&\text { s.t. }  \frac{\partial}{\partial t} V(z,t_\text{return}) + \min_{\eta\in E}
%\min_{d \in \mathcal{D}_{\mathrm{tv}}} 
\nabla_{z} V(z, t_\text{return})^\top \hat{f}(z,u,\eta)\geq- \alpha(V(z, t_\text{return})),
\end{aligned}
\end{equation}

with $E = \mathcal{D}_{\text{tv}}(\overline{D_{\tilde{f}}d}, t_\text{return})$. 
This safety filter ensures the system is \textit{realistic} about the present (using the current disturbance $\bar{d}$) and \textit{pessimistic} about the future (assuming the worst-case rate of change to persist), which strongly reduces conservativeness while still providing a margin of safety in face of varying disturbances. 
Algorithm~\ref{alg:cap} and Fig.~\ref{fig:method_time_to_space} provide an overview of the overall \algname{} framework. 

\vspace{-0.1cm}
\section{Experimental Results} \label{sec:result}
We evaluate \algname{} extensively in both simulation and hardware. 
Our experiments highlight how our disturbance reparameterization improves safety and goal-reaching performance compared to baselines that do not explicitly account for disturbance variations. 

\textbf{Learning Value Functions}
We learn value functions using Deepreach~\cite{BansalTomlin2021}, a self-supervised PINN, across a wide range of disturbances rates $\dot{d}$, for our method, and disturbance magnitudes $|d|$ for the baselines. 
Unlike the baselines, \algname{} solves a reach-avoid problem with the additional requirement that the reach set be control invariant. 
This requires formulating the problem as a reach-set avoid-tube problem~\cite{Begzadic2025BackTB}, which we find empirically easier to learn than the standard avoid tube formulation (see Appendix~\ref{sec:training_details} for training details and visualizations of the learned safe sets). 

\begin{table}[b]
\vspace{-0.7cm}
\caption{Comparison of our approach against baselines using HJR and Deepreach based value functions. Metrics are generated over $50$ trajectories across randomized wind fields, with $10$ goals and $1000$ control steps each. $\%$ Safety Violations indicates the $\%$ of failed trajectories, Mean Goal Distance reflects the average minimum distance to each goal, and Mean Trajectory Length refers to the mean trajectory length before failure with a max of $1000$. The table highlights how \algname{} provides the best balance between safety and performance.}
\centering
\begin{tabular}{l@{\hskip -0.1in}ccc}
\toprule
\text{Approach} & \text{\% Safety Violations $\downarrow$} & \text{Mean Goal Distance $\downarrow$} & \text{Mean Trajectory Length $\uparrow$} \\
\midrule
HJR Naive & 96\% & 1.09 & 415 \\
HJR Naive Worstcase & 0\% & 1.95 & 1000 \\
DeepReach Naive & 90\% & 1.19 & 472 \\
\textbf{DeepReach Ours} & 34\% & 1.02 & 781 \\
\textbf{HJR Ours} & 2\% & 0.78 & 993 \\
\bottomrule
\end{tabular}
\label{tab:domain_comparison}
\end{table}

\begin{figure}[t]
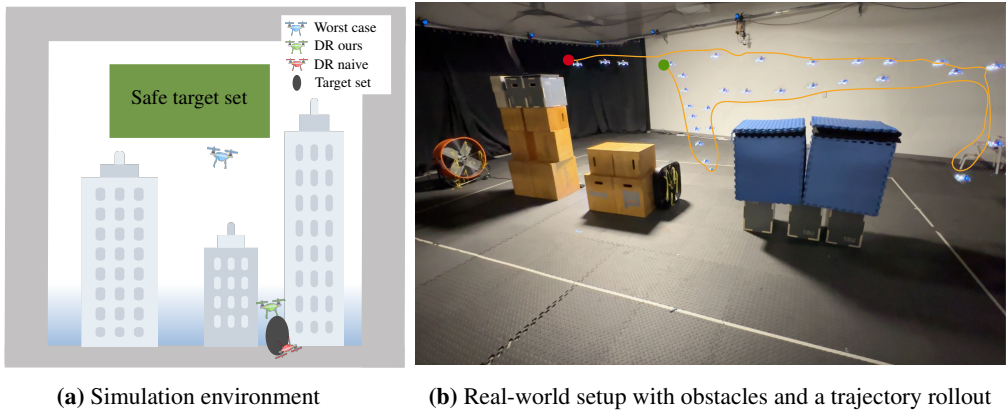

    \centering
    \begin{subfigure}{0.4\linewidth}
        \centering
        \scalebox{0.78}{\input{Tikz-related-figs/figure_3.tex}}
        \caption{Simulation environment}
        \label{fig:sim_env}
    \end{subfigure}
    \begin{subfigure}{0.58\linewidth}
        \centering
        \resizebox{\linewidth}{!}{\input{Tikz-related-figs/figure_4_ours_dist_traj}}
        \caption{Real-world setup with obstacles and a trajectory rollout}
        \label{fig:experiment-figure}
    \end{subfigure}
    \caption{
    Fig.~\ref{fig:sim_env} illustrates the cityscape simulation environment along with simulation results. 
    The blue color gradient indicates increasing wind disturbance in the urban canyons. 
    The drones in the figure illustrate the performance of our method and the baselines when attempting to go to the goal region denoted by the dark gray oval.
    Fig.~\ref{fig:experiment-figure} illustrates our real world experiment setup using crazyflies (with motion capture), mimicing Fig.~\ref{fig:sim_env}. A single successful \algname rollout is overlayed.
    }
    \label{fig:combined_experiment}
    \vspace{-0.5cm}
\end{figure}

\textbf{Simulation Experiments}
We validate our approach in a planar drone environment with state $[p_x, p_z, v_x, v_z]$, where the drone navigates the $x$-$z$ plane through a simulated cityscape, depicted in Fig.~\ref{fig:sim_env}. 
This setup allows us to validate the proposed reparameterization in \algname{} with ground truth value functions computed via dynamic programming-based Hamilton Jacobi Reachability (HJR). 
We conduct extensive evaluations of \algname{} using both learned and HJR-computed value functions. 
In our experiments we compare the following controllers: 
i) \textbf{HJR Ours: } \algname{} with an ensemble of HJR value functions with different fixed disturbance rates. 
ii) \textbf{Deepreach (DR) Ours: } \algname{} with a learned value function parameterized by disturbance rate. 
iii) \textbf{HJR Naive: } An ensemble of HJR value functions each computed with different  disturbance bounds. 
iv) \textbf{HJR Naive Worst-Case: } HJR value function assuming the worst-case disturbances $\mathcal{D}_{\max}$ everywhere. 
v) \textbf{DR Naive: } Learned value function parameterized by disturbance bounds~\cite{Borquez2022ParameterConditionedRS}.

In our simulated cityscape environment, spatial variations in wind intensity are modeled as an exponential function of altitude, $z$, intensifying towards the bottom of an urban canyon. 
Disturbances act on both position and velocity states in the dynamics.
The parameters characterizing the wind are randomized for every trajectory. 
The safe control-invariant reach set corresponds to a fly-over zone above all buildings. 
This scenario reflects real-world inspection and delivery tasks, where drones must traverse high-disturbance regions near structures while reaching multiple sequential goals. 
Additional modeling details are in Appendix~\ref{sec:exp_design}. 

All compared methods use a recency-based disturbance estimation strategy with $H=1$, where they receive disturbance measurements at a rate of $4$Hz. 
The drone controller is run at a faster rate of $40$Hz.
Results are summarized in Table~\ref{tab:domain_comparison}. 
Our method consistently achieves lower safety violation rates and better goal-reaching performance. 
The higher violation rate for DR Ours is largely due to inaccuracies in the learned value function, with many violations occurring outside the wind-affected areas. 
Fig. \ref{fig:sim_env} also shows representative trajectories comparing DR ours (green drone) to DR Naive (red drone) and HJR Worst Case baseline (blue drone).
HJR Worst-Case is overly conservative and often never leaves the initial safe region; DR Naive fails to adapt to changing disturbances and crashes; in contrast, DR Ours safely navigates to the goal. 
Traces of all simulated trajectories are visualized in Fig.~\ref{fig:ours_alltraj} and Fig.~\ref{fig:baseline_alltraj} in Section~\ref{sec:exp_design} of the Appendix.
\begin{figure}[t!]
\centering
\begin{subfigure}[t]{0.5\linewidth}
  \centering
  \resizebox{\linewidth}{!}{\includegraphics{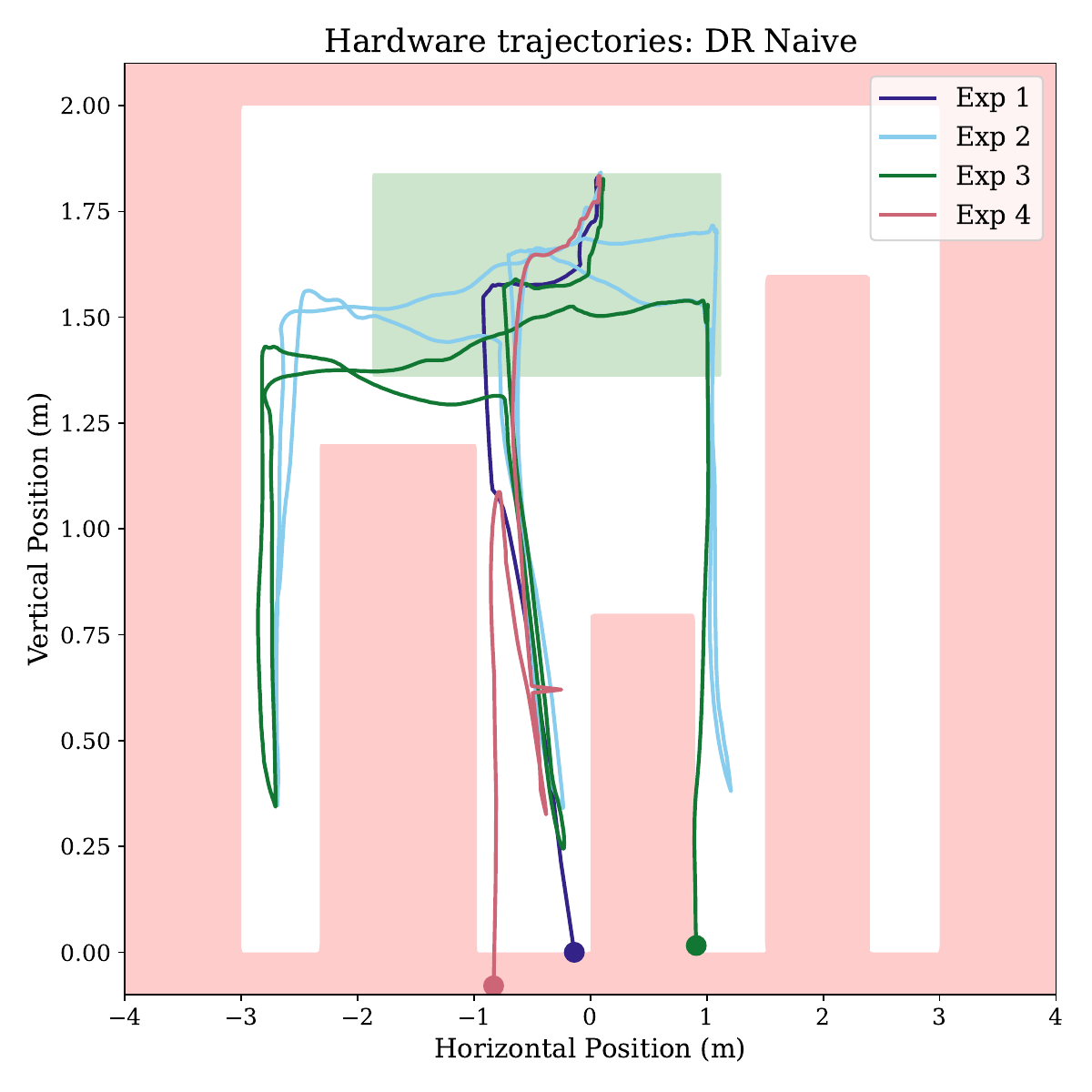}}
\end{subfigure}%
\begin{subfigure}[t]{0.5\linewidth}
  \centering
  \resizebox{\linewidth}{!}{\includegraphics{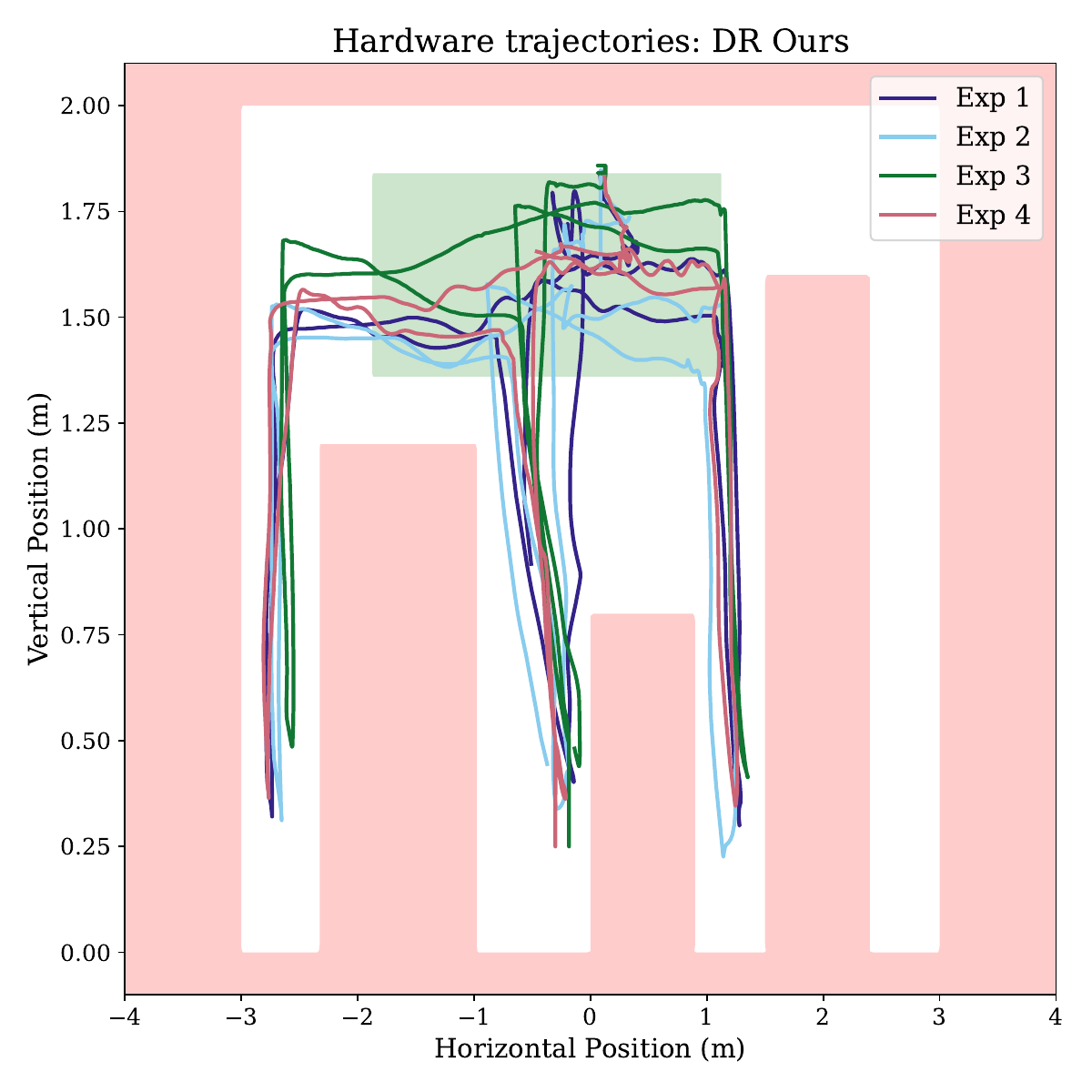}}
\end{subfigure}
\caption{Trajectories of real-world hardware experiments, comparing \textbf{Naive} (left) and \textbf{Ours} (right). 
\algname successfully accounts for the disturbance increase as we descend the urban canyons. In contrast, \textbf{Naive} fails to adequately adapt, leading to a crash in 4 out of 5 rollouts.}
\label{fig:realworld}
\vspace{-0.5cm}
\end{figure}

\textbf{Hardware Experiments: }
We validate our approach on hardware using Crazyflie drones and an OptiTrack motion capture system. 
To capture realistic drone behavior, we use a more complex 6D drone model with state $[p_x,v_{x}, \theta_{x}, \omega_{x}, p_z, v_{z}]$ where $\theta_{x}$ is the pitch and $\omega_{x}$ is the angular velocity and control inputs of desired pitch and thrust. 
Due to the high dimensionality (8D including parametric dimensions), traditional HJR methods (and therefore online fine-tuning methods~\cite{Fisac2017AGS, Herbert2021ScalableLO} are computationally intractable, motivating the need for learned value functions which capture changes in disturbance bounds. 

We recreate the simulated urban environment using stacked boxes, as shown in  Fig.~\ref{fig:experiment-figure}. 
Given the Crazyflies' sensitivity to non-uniform wind gusts, we instead spoofed disturbances for our hardware experiments by adding them directly to the state observations from the OptiTrack. 
We run the drone with a control rate of $50$Hz, while disturbance sampling is run at $5$Hz. 
We compare \algname{} (DR Ours) against DR Naive~\cite{Borquez2022ParameterConditionedRS} which naively switches between fixed disturbance bounds.  
As in simulation, our method preserves safety over $5$ trials, while the naively switching baseline crashes ($4/5$ trials) as the drone fails to adequately adapt to the changing disturbances, see Fig.~\ref{fig:realworld}. 
The results underscore the necessity of our approach for safe navigation in real-world disturbance varying environments. 
Appendix~\ref{sec:exp_design} provides further details on the experimental setup and results.

\vspace{-0.2cm}
\section{Conclusion}
\label{sec:conclusion}
\vspace{-0.1cm}
This paper introduced \algname{}, a novel framework for enabling learning value functions offline for deployment in environments with unknown, spatially-varying disturbances. Specifically, by reparameterizing spatial disturbances into temporally varying disturbances, our method leverages the scalability of offline learning, while providing the adaptability required for real-world operation. 
We validated \algname{} through extensive simulations and real-world experiments, demonstrating improved safety performance compared to baselines that do not explicitly account for changing disturbances in their value functions. 
Reducing conservativeness through tighter online learning integration and improving robustness to model uncertainty are key directions to broaden the applicability of our framework in more complex environments.
\newpage
\section{Limitations}
\label{sec:limitations}
\algname{} has several limitations, which limit its generalizability and its readiness for deployment. First, as we parameterize spatially varying disturbances (often in 3D) using time (a scalar), the method is inherently conservative. 
Concretely, \algname{} sets its ``time to return'' based on the worst combination of the worst-case receding horizon estimated disturbance magnitude and directional derivative values, which results in a more restrictive safety filter than if all dimensions of the disturbance could be captured independently.
In addition, \algname{} assumes we can accurately characterize the disturbance bound and its directional flow for any state (or assumes a conservative worst-case disturbance rate in the entire state space). 
Lastly, our implementation and associated results, which are based on Deepreach~\cite{BansalTomlin2021}, require an accurate dynamics model. 
Nontheless, \algname{} can be extended to reinforcement learning-based frameworks that do not require a dynamics model. 
\clearpage
\bibliography{main} 

\begin{thebibliography}{26}
\providecommand{\natexlab}[1]{#1}
\providecommand{\url}[1]{\texttt{#1}}
\expandafter\ifx\csname urlstyle\endcsname\relax
  \providecommand{\doi}[1]{doi: #1}\else
  \providecommand{\doi}{doi: \begingroup \urlstyle{rm}\Url}\fi

\bibitem[Hsu et~al.(2023)Hsu, Hu, and Fisac]{Hsu2023TheSF}
K.-C. Hsu, H.~Hu, and J.~F. Fisac.
\newblock {The Safety Filter: A Unified View of Safety-Critical Control in Autonomous Systems}.
\newblock \emph{Annu. Rev. Control. Robotics Auton. Syst.}, 7, 2023.

\bibitem[Ames et~al.(2017)Ames, Xu, Grizzle, and Tabuada]{Ames2017ControlBF}
A.~D. Ames, X.~Xu, J.~W. Grizzle, and P.~Tabuada.
\newblock {Control Barrier Function} based quadratic programs for safety critical systems.
\newblock In \emph{{IEEE Transactions on Automatic Control}}, volume~62, pages 3861--3876, 2017.

\bibitem[Bansal et~al.(2017)Bansal, Chen, Herbert, and Tomlin]{Bansal2017HamiltonJacobiRA}
S.~Bansal, M.~Chen, S.~L. Herbert, and C.~J. Tomlin.
\newblock {Hamilton-Jacobi} reachability: A brief overview and recent advances.
\newblock In \emph{{Proc.\ IEEE Conf.\ on Decision and Control}}, 2017.

\bibitem[Choi et~al.(2021)Choi, Lee, Sreenath, Tomlin, and Herbert]{ChoiLeeEtAl2021}
J.~J. Choi, D.~Lee, K.~Sreenath, C.~J. Tomlin, and S.~L. Herbert.
\newblock Robust {Control Barrier-Value Functions} for safety-critical control.
\newblock In \emph{{Proc.\ IEEE Conf.\ on Decision and Control}}, 2021.

\bibitem[Tonkens and Herbert(2022)]{TonkensHerbert2022}
S.~Tonkens and S.~Herbert.
\newblock Refining {Control Barrier Functions} through {Hamilton-Jacobi} reachability.
\newblock In \emph{{IEEE/RSJ Int.\ Conf.\ on Intelligent Robots \& Systems}}, 2022.

\bibitem[Begzadi\'c et~al.(2025)Begzadi\'c, Shinde, Tonkens, Hirsch, Ugalde, Yip, Cort\'es, and Herbert]{Begzadic2025BackTB}
A.~Begzadi\'c, N.~Shinde, S.~Tonkens, D.~Hirsch, K.~Ugalde, M.~C. Yip, J.~Cort\'es, and S.~Herbert.
\newblock {Back to Base: Towards Hands-Off Learning via Safe Resets with Reach-Avoid Safety Filters}.
\newblock \emph{ArXiv}, abs/2501.02620, 2025.

\bibitem[Bansal and Tomlin(2021)]{BansalTomlin2021}
S.~Bansal and C.~J. Tomlin.
\newblock {DeepReach: A Deep Learning Approach to High-Dimensional Reachability}.
\newblock In \emph{{Proc.\ IEEE Conf.\ on Robotics and Automation}}, 2021.

\bibitem[Hsu et~al.(2021)Hsu, Rubies-Royo, Tomlin, and Fisac]{HsuRubies-RoyoEtAl2021}
K.-C. Hsu, V.~Rubies-Royo, C.~J. Tomlin, and J.~F. Fisac.
\newblock {Safety and Liveness Guarantees through Reach-Avoid Reinforcement Learning}.
\newblock In \emph{{Robotics: Science and Systems}}, 2021.

\bibitem[So et~al.(2024)So, Serlin, Mann, Gonzales, Rutledge, Roy, and Fan]{SoSerlinEtAl2024}
O.~So, Z.~Serlin, M.~Mann, J.~Gonzales, K.~Rutledge, N.~Roy, and C.~Fan.
\newblock {How to Train Your Neural {Control Barrier Function}: Learning Safety Filters for Complex Input-Constrained Systems}.
\newblock In \emph{{Proc.\ IEEE Conf.\ on Robotics and Automation}}, 2024.

\bibitem[Nguyen et~al.(2024)Nguyen, Hsu, Yu, Tan, and Fisac]{Nguyen2024GameplayFR}
D.~P. Nguyen, K.-C. Hsu, W.~Yu, J.~Tan, and J.~F. Fisac.
\newblock {Gameplay Filters: Robust Zero-Shot Safety through Adversarial Imagination}.
\newblock In \emph{{Conf.\ on Robot Learning}}, 2024.

\bibitem[Hunter et~al.(1990)Hunter, Watson, and Johnson]{HUNTER1990315}
L.~Hunter, I.~Watson, and G.~Johnson.
\newblock Modelling air flow regimes in urban canyons.
\newblock \emph{Energy and Buildings}, 15\penalty0 (3):\penalty0 315--324, 1990.

\bibitem[Achermann et~al.(2019)Achermann, Lawrance, Ranftl, Dosovitskiy, Chung, and Siegwart]{AchermannLawranceEtAl2019}
F.~Achermann, N.~R.~J. Lawrance, R.~Ranftl, A.~Dosovitskiy, J.~J. Chung, and R.~Siegwart.
\newblock Learning to predict the wind for safe aerial vehicle planning.
\newblock In \emph{{Proc.\ IEEE Conf.\ on Robotics and Automation}}, 2019.

\bibitem[Brunke et~al.(2022)Brunke, Greeff, Hall, Yuan, Zhou, Panerati, and Schoellig]{brunke2022safe}
L.~Brunke, M.~Greeff, A.~W. Hall, Z.~Yuan, S.~Zhou, J.~Panerati, and A.~P. Schoellig.
\newblock {Safe Learning in Robotics: From Learning-Based Control to Safe Reinforcement Learning}.
\newblock \emph{{Annual Review of Control, Robotics, and Autonomous Systems}}, 5:\penalty0 411--444, 2022.

\bibitem[Wabersich et~al.(2023)Wabersich, Taylor, Choi, Sreenath, Tomlin, Ames, and Zeilinger]{Wabersich2023DataDrivenSF}
K.~P. Wabersich, A.~J. Taylor, J.~J. Choi, K.~Sreenath, C.~J. Tomlin, A.~Ames, and M.~N. Zeilinger.
\newblock {Data-Driven Safety Filters: Hamilton-Jacobi Reachability, Control Barrier Functions, and Predictive Methods for Uncertain Systems}.
\newblock \emph{IEEE Control Systems}, 43:\penalty0 137--177, 2023.

\bibitem[Fisac et~al.(2019)Fisac, Lugovoy, {Rubies-Royo}, Ghosh, and Tomlin]{FisacLugovoyEtAl2019}
J.~F. Fisac, N.~F. Lugovoy, V.~{Rubies-Royo}, S.~Ghosh, and C.~J. Tomlin.
\newblock Bridging {Hamilton-Jacobi} safety analysis and reinforcement learning.
\newblock In \emph{{Proc.\ IEEE Conf.\ on Robotics and Automation}}, 2019.

\bibitem[Borquez et~al.(2022)Borquez, Nakamura, and Bansal]{Borquez2022ParameterConditionedRS}
J.~Borquez, K.~Nakamura, and S.~Bansal.
\newblock {Parameter-Conditioned Reachable Sets for Updating Safety Assurances Online}.
\newblock In \emph{{Proc.\ IEEE Conf.\ on Robotics and Automation}}, 2022.

\bibitem[Tayal and Kolathaya(2023)]{Tayal2023ControlBF}
M.~Tayal and S.~N.~Y. Kolathaya.
\newblock {Control Barrier Functions in Dynamic UAVs for Kinematic Obstacle Avoidance: A Collision Cone Approach}.
\newblock In \emph{{American Control Conference}}, 2023.

\bibitem[Yu et~al.(2024)Yu, Yu, Naddaf-Sh, Upadhyay, Gao, and Fan]{Yu2024EfficientMP}
M.~Yu, C.~Yu, M.-M. Naddaf-Sh, D.~Upadhyay, S.~Gao, and C.~Fan.
\newblock {Efficient Motion Planning for Manipulators with Control Barrier Function-Induced Neural Controller}.
\newblock In \emph{{Proc.\ IEEE Conf.\ on Robotics and Automation}}, 2024.

\bibitem[Grandia et~al.(2020)Grandia, Taylor, Ames, and Hutter]{Grandia2020MultiLayeredSF}
R.~Grandia, A.~J. Taylor, A.~Ames, and M.~Hutter.
\newblock {Multi-Layered Safety for Legged Robots via Control Barrier Functions and Model Predictive Control}.
\newblock In \emph{{Proc.\ IEEE Conf.\ on Robotics and Automation}}, 2020.

\bibitem[Robey et~al.(2020)Robey, Hu, Lindemann, Zhang, Dimarogonas, Tu, and Matni]{Robey2020LearningCB}
A.~Robey, H.~Hu, L.~Lindemann, H.~Zhang, D.~V. Dimarogonas, S.~Tu, and N.~Matni.
\newblock {Learning Control Barrier Functions from Expert Demonstrations}.
\newblock In \emph{{Proc.\ IEEE Conf.\ on Decision and Control}}, 2020.

\bibitem[Tayal et~al.(2024)Tayal, Zhang, Jagtap, Clark, and Kolathaya]{Tayal2024LearningAF}
M.~Tayal, H.~Zhang, P.~Jagtap, A.~Clark, and S.~N.~Y. Kolathaya.
\newblock {Learning a Formally Verified Control Barrier Function in Stochastic Environment}.
\newblock In \emph{{Proc.\ IEEE Conf.\ on Decision and Control}}, 2024.

\bibitem[Manda et~al.(2025)Manda, Chen, and Fazlyab]{pmlr-v270-manda25a}
L.~Manda, S.~Chen, and M.~Fazlyab.
\newblock {Learning Performance-oriented Control Barrier Functions Under Complex Safety Constraints and Limited Actuation}.
\newblock In \emph{{Conf.\ on Robot Learning}}, 2025.

\bibitem[Jeong et~al.(2024)Jeong, Gong, Bansal, and Herbert]{Jeong2024ParameterizedFA}
H.~J. Jeong, Z.~Gong, S.~Bansal, and S.~L. Herbert.
\newblock {Parameterized Fast and Safe Tracking (FaSTrack) using Deepreach}.
\newblock In \emph{{Learning for Dynamics \& Control}}, 2024.

\bibitem[Fisac et~al.(2017)Fisac, Akametalu, Zeilinger, Kaynama, Gillula, and Tomlin]{Fisac2017AGS}
J.~F. Fisac, A.~K. Akametalu, M.~N. Zeilinger, S.~Kaynama, J.~H. Gillula, and C.~J. Tomlin.
\newblock A general safety framework for learning-based control in uncertain robotic systems.
\newblock In \emph{{IEEE Transactions on Automatic Control}}, volume~64, pages 2737--2752, 2017.

\bibitem[Herbert et~al.(2021)Herbert, Choi, Qazi, Gibson, Sreenath, and Tomlin]{Herbert2021ScalableLO}
S.~L. Herbert, J.~J. Choi, S.~Qazi, M.~T. Gibson, K.~Sreenath, and C.~J. Tomlin.
\newblock {Scalable Learning of Safety Guarantees for Autonomous Systems using Hamilton-Jacobi Reachability}.
\newblock In \emph{{Proc.\ IEEE Conf.\ on Robotics and Automation}}, 2021.

\bibitem[Rasmussen and Williams(2006)]{Rasmussen2006}
C.~Rasmussen and C.~Williams.
\newblock \emph{{Gaussian Processes for Machine Learning}}.
\newblock MIT Press, Cambridge, 2006.

\end{thebibliography}
\newpage
\appendix
\section{Theoretical Background} \label{sec:appendix_th_background}

\subsection{Computing the Reachability Problem}
Recall that Hamilton–Jacobi reachability (HJR) formulates the problem as a differential game between two players: the control input $u$, which seeks to maximize the reward, and the disturbance $d$, which acts adversarially to minimize it. Then, the value function of this game is defined as 
\begin{equation} \label{eq:value_fun}
    V(x,t)= \min_{\mathbf{d} \in \mathbb{D}(t)}\max_{\mathbf{u}\in\mathbb{U}(t)} r_{\text{RA}}(x,t,\bm{u}, \bm{d}) = \min_{\mathbf{d} \in \mathbb{D}(t)}\max_{\mathbf{u}\in\mathbb{U}(t)} \max_{\tau \in [t,0]}\min\{l(\bm{x}_{x,t}^{\bm{u},\bm{d}}(\tau)), \min_{s \in [t,\tau]}g(\bm{x}_{x,t}^{\bm{u}, \bm{d}}(s))\}.
\end{equation}
In general, solving the value function optimization problem in \eqref{eq:value_fun} is non-convex and therefore challenging. 
However, using dynamic programming (backwards in time), the value function $V$ is the unique viscosity solution of the following Hamilton-Jacobi-Isaacs Variational Inequality (HJI-VI) 
\begin{equation}
    0 = \min\left\{g(x) - V(x,t), \max\left\{l(x) - V(x,t),  \frac{\partial}{\partial t} V(x,t) + H(\nabla V(x,t), x)\right\}\right\},
\end{equation}
with the Hamiltonian defined by $H(\lambda,x)~=~\max\limits_{u \in \mathcal{U}}\min\limits_{d \in \mathcal{D}}\lambda^\top f(x,u,d)$ and terminal cost $V(x,0)=\min\{l(x), g(x)\}$. The gradients of this function enable the computation of the optimal safety control  $u^*(x,t)$ such that 
\begin{equation}
    u^*(x,t) = \arg\max_{u\in\mathcal{U}} \min_{d\in\mathcal{D}} \nabla V(x,t)^\top \tilde{f}(x,u,d).
\end{equation}
Value functions can be computed effectively via dynamic programming in low-dimensional systems by discretizing the state space to form a high-resolution grid~\cite{Bansal2017HamiltonJacobiRA}. However, the reliance on grid-based discretization makes such methods scale exponentially with the system's state dimension, making these methods unsuitable for systems with more than $5-6$ state variables.
\subsection{Blending Reachability Analysis with Control Barrier Functions}
Constructing valid CBFs for complex systems with input bounds and disturbances is often challenging, especially when the safe set is difficult to characterize analytically. By leveraging reachability analysis, one can systematically synthesize CBF-like safety filters to ensure safety. For instance, HJR reach-avoid problems, with value function $h_{v}$, can be integrated with CBFs, leading to the following definition.
\begin{definition}(\textbf{Viscosity-Based Control Barrier Function~\cite{Begzadic2025BackTB}}) \label{def:ra-cbvf}
Consider a continuous function $h_v: \mathbb{R}^n \times (-\infty, 0] \to \mathbb{R}$, and for each $t \le 0$, let $\mathcal{C}_v(t) = \{ x \in \mathbb{R}^n \mid h_v(x,t) \ge 0 \}$. Then $h_v$ is a viscosity-based control barrier function (VB-CBF) for system \eqref{eq:orig_sys} on $\mathcal{C}_v(\cdot)$ if there exists an extended class $\mathcal{K}$ function $\alpha$ such that for all $t < 0$ and all $x \in \mathcal{C}_v(t)$, the inequality
       $\frac{\partial}{\partial t} h_v(x,t) + \max_{u \in \mathcal{U}} \min_{d \in \mathcal{D}} \nabla_x h_v(x, t)^\top \tilde{f}(x, u, d)  \geq - \alpha(h_v(x, t))$
holds in a viscosity sense.
\end{definition}
Given a nominal control policy $u_\text{nom}$ that may violate input or safety constraints, viscosity-based control barrier functions enable minimal modification of the nominal input via a quadratic program. The resulting minimally invasive safety filter not only ensures constraint satisfaction but also guides the system away from unsafe regions and back toward the desired target set.

\section{Learning-Based Reachability Analysis}\label{sec:appendix_learning_ra}
Several learning-based approaches have been developed to approximate the reach-avoid reachability value function. Although our framework is compatible with a broad range of methods (including self-supervised learning~\cite{BansalTomlin2021} and reinforcement learning~\cite{HsuRubies-RoyoEtAl2021}), we implement our method using Deepreach:

\textbf{Self-Supervised Learning of Reachability Value Functions}
To avoid solving HJI-VI with grid‑based discretization, we employ a Physics‑Informed Neural Network (PINN) to learn the value function used in minimally invasive safety filters. 
In particular, we leverage the DeepReach framework to approximate the safety value function by employing a sinusoidal deep neural network architecture~\cite{BansalTomlin2021}. Consequently, the computational and memory demands of training depend on the intrinsic complexity of the value‑function approximation rather than on the grid resolution. For a reach-avoid problem, the loss function $\mathcal{L}$ used for training DeepReach is given by
\begin{align}\label{eq:loss_and_hamiltonian}
\mathcal{L}(\theta) &= \mathbb{E}_{z, t} \left[
\Bigg\Vert \min\Bigg\{g(x) - V_\theta(z,t),
\max\bigg\{
l(x) - V_\theta(z, t),\;
\frac{\partial V_\theta}{ \partial t} + H(\nabla_z V_\theta, z)
\bigg\}\Bigg\}
\Bigg\Vert \right], 
\end{align}

where $V_\theta(z,t)= \min\{l(x),g(x)\} - t \cdot \text{NN}_{\theta}(z, t)$, with $z=[x, p]$ the joint state, $x$ the model state and $p$ the parameterized state, which are discussed in more detail in Appendix~\ref{sec:exp_design}.

The loss function~\eqref{eq:loss_and_hamiltonian} involves nested $\min$ and $\max$ operations; 
These operations induce non-smooth behavior in the loss function 
This poses a challenge for neural networks, as learning relies on backpropagation and smooth gradient flow to update the model parameters effectively. 
Moreover, the smoothness of the learned value function is critical, as our safety filter relies directly on the gradient of the value function to ensure safe control inputs (motivating~\cite{BansalTomlin2021}'s use of sinusoidal activation functions, which we also employ). 

To improve the learning process, we leverage a key observation relevant to our application; The target set characterizing the reach-avoid tube problem is control invariant. 
That is, once the drone reaches the target set, it can remain within this set and maintain safety under any level of disturbances. 
This property allows reformulating the learning objective by shifting the focus from learning a value function to reach a target while avoiding unsafe regions over time, to instead learning or defining a control-invariant set and learning a value function to avoid unsafe regions over time, while only implictly encoding the reaching of the target in the boundary condition.
Hence, the modified reach-avoid control-invariant loss function for DeepReach is given by 
\begin{equation}\label{eq:loss_and_hamiltonian_raci}
\mathcal{L}(\theta) = \mathbb{E}_{z, t} \left[
\bigg\Vert \min\bigg\{g(x) - V_\theta(z,t),
\frac{\partial V_\theta}{ \partial t} + H(\nabla_z V_\theta, z)
\bigg\}
\bigg\Vert \right],
\end{equation}
where $V_\theta(z,t)= \min\{l(x),g(x)\} - t \cdot \text{NN}_{\theta}(z, t)$, like before.
We empirically observe smoother gradients and a better-learned solution employing~\eqref{eq:loss_and_hamiltonian_raci}.

\section{DeepReach Training Details}\label{sec:training_details}
To evaluate our method and relevant baselines, we adopt a modified, parameterized version of DeepReach (DR) to learn the value function under different environmental conditions. 

For the \textbf{DR Naive} baseline, the parameterized inputs correspond to the disturbance magnitude (applied to position and velocity in $x$ and $z$ directions in the 4D model and only to the velocity components in the $x$ and $z$ directions in the 6D model). 

For \textbf{DR ours}, the parameterized inputs correspond to the disturbance rate, which defines how the disturbance magnitude varies with the 
% change in the 
environment. 
This includes disturbance rates over both position and velocity disturbance magnitudes in the $x$ and $z$ directions in the 4D model and disturbance rates for the velocity disturbance magnitude in the $x$ and $z$ directions in the 6D model.  

\textbf{Parameterized Value Function Using DeepReach. }To incorporate these parameters, we modify DR following the approach in~\cite{Borquez2022ParameterConditionedRS} to account for the environmental conditions, such as disturbance bounds or rates, as part of the input space. We augment the system with the disturbance rate $\dot{d}$ as in \eqref{eq:sys_new_dotd} and compute the backward reachable tube for the resulting parameterized system using DR. Accordingly, the neural network takes as input the state $z=[x, \dot{d}]$ and time $t$, and outputs the corresponding value function $V_\theta(z, t)$, where $\theta$ denotes the network parameters. For the baselines we consider the joint state $z=[x, d]$, with $d$ the disturbance magnitude. 
We generate training inputs using uniform sampling over both the state and parameter dimensions, covering the desired range of environmental conditions. 
The model is trained with default Deepreach settings, most importantly a batch size of $65$k states, over $100$k steps with a learning rate $\eta = 2e^{-5}$. 
It uses the default parameters of the Deepreach repository \url{https://github.com/smlbansal/deepreach/tree/public_release}. For the 4D simulation model, we use~\eqref{eq:loss_and_hamiltonian} for the loss, which had adequate performance for this setting.
However, for the 6D hardware experiments, we adopt the reach-avoid control-invariant loss~\eqref{eq:loss_and_hamiltonian_raci} which leads to a better solution.

\textbf{Training Challenges. }For the DR baselines, we initially aimed to learn a time-invariant always-avoid value function for the environment with varying maximum disturbance bounds. However, due to convergence difficulties, this approach did not yield a sufficiently performant solution. To ensure a fair comparison, we instead trained DR naive on a reach-avoid formulation toward the same target set as our method, while instead parameterizing over the disturbance bounds. This formulation enabled successful training and produced reliable value functions. During deployment, we evaluated the value function solely at the final time point, effectively considering an avoid-only value function, which allowed us to construct a time-invariant, minimally invasive safety filter.

\begin{figure}[t!]
\centering
\begin{subfigure}[t]{0.33\linewidth}
  \centering
  \resizebox{\linewidth}{!}{\includegraphics{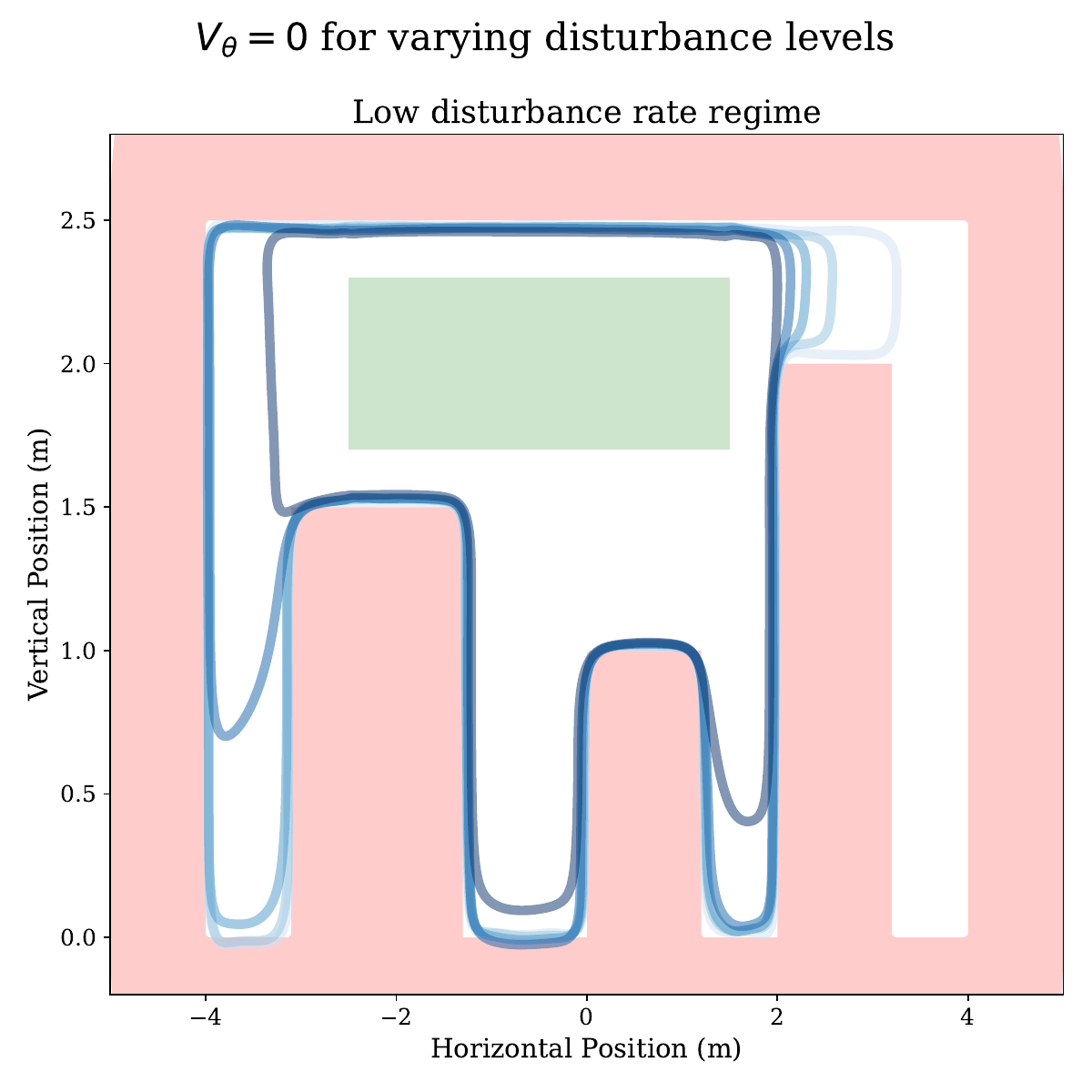}}
\end{subfigure}%
\begin{subfigure}[t]{0.33\linewidth}
  \centering
  \resizebox{\linewidth}{!}{\includegraphics{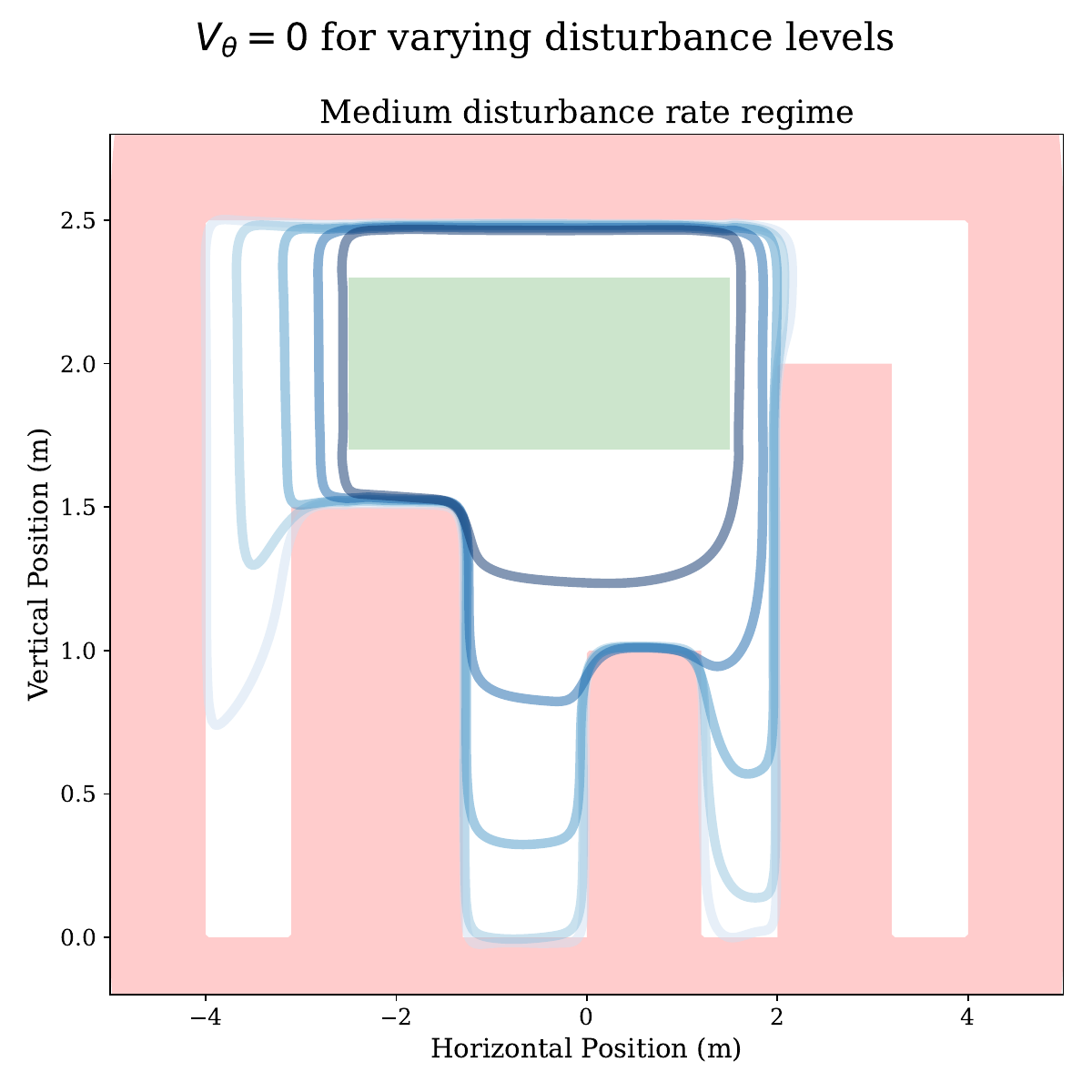}}
\end{subfigure}
\begin{subfigure}[t]{0.33\linewidth}
  \centering
  \resizebox{\linewidth}{!}{\includegraphics{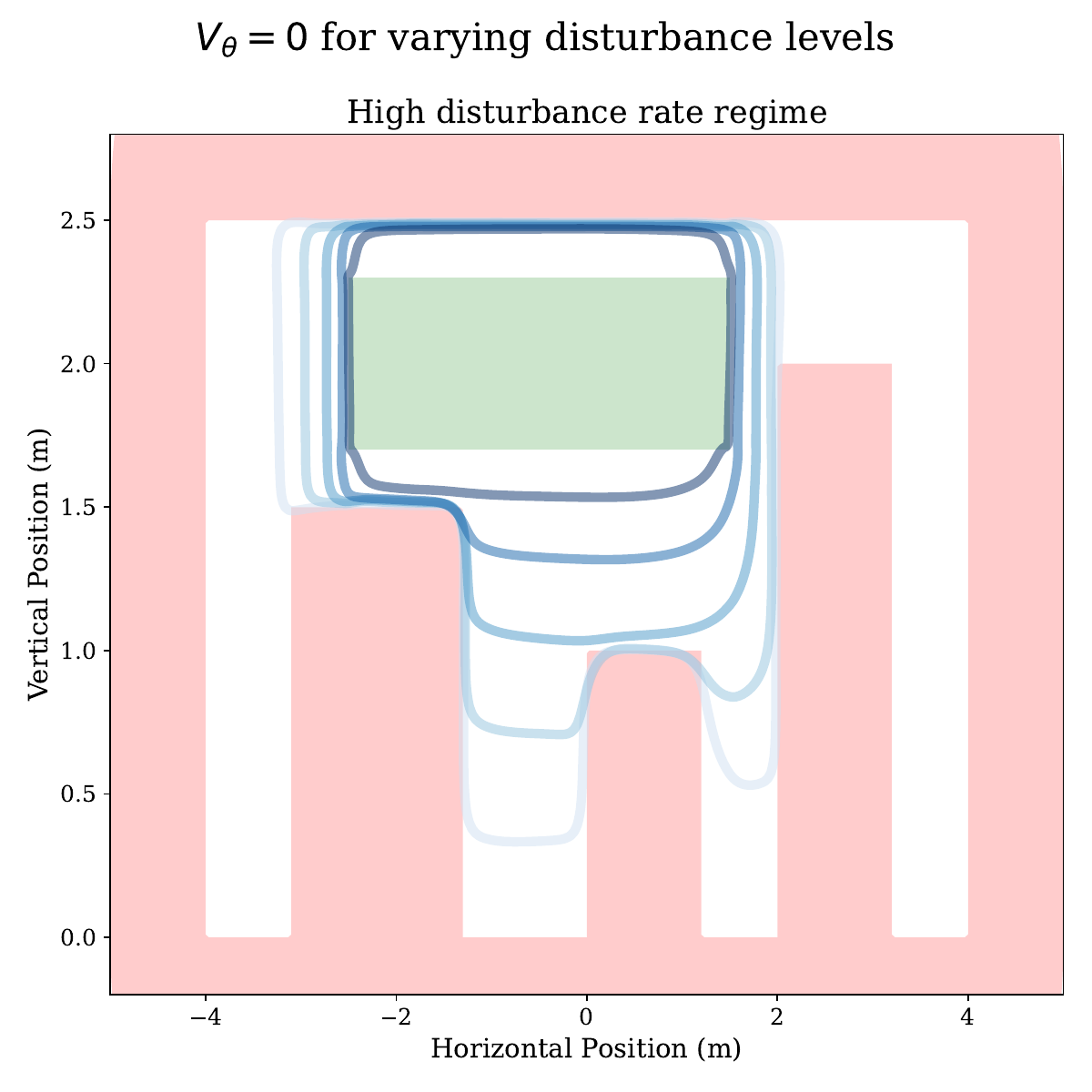}}
\end{subfigure}
\caption{The 0-level set of the learned value function for a fixed disturbance rate $\dot{d}$ for increasing levels (light to dark blue) of disturbance $d$. Left-to-right visualizes a low fixed disturbance rate, a medium disturbance rate, and a high disturbance rate respectively. This is evaluated for the $v_x=0, v_z=0$ slice. $\dot{d}$ is encodes through the parameterized state, whereas $d$ is encoded through the time slice of the value function with $t=(d_\text{max} - d)/\dot{d}$}
\label{fig:disturbance_levelsets}
\end{figure}

\begin{figure}[t!]
\centering
\begin{subfigure}[t]{0.33\linewidth}
  \centering
  \resizebox{\linewidth}{!}{\includegraphics{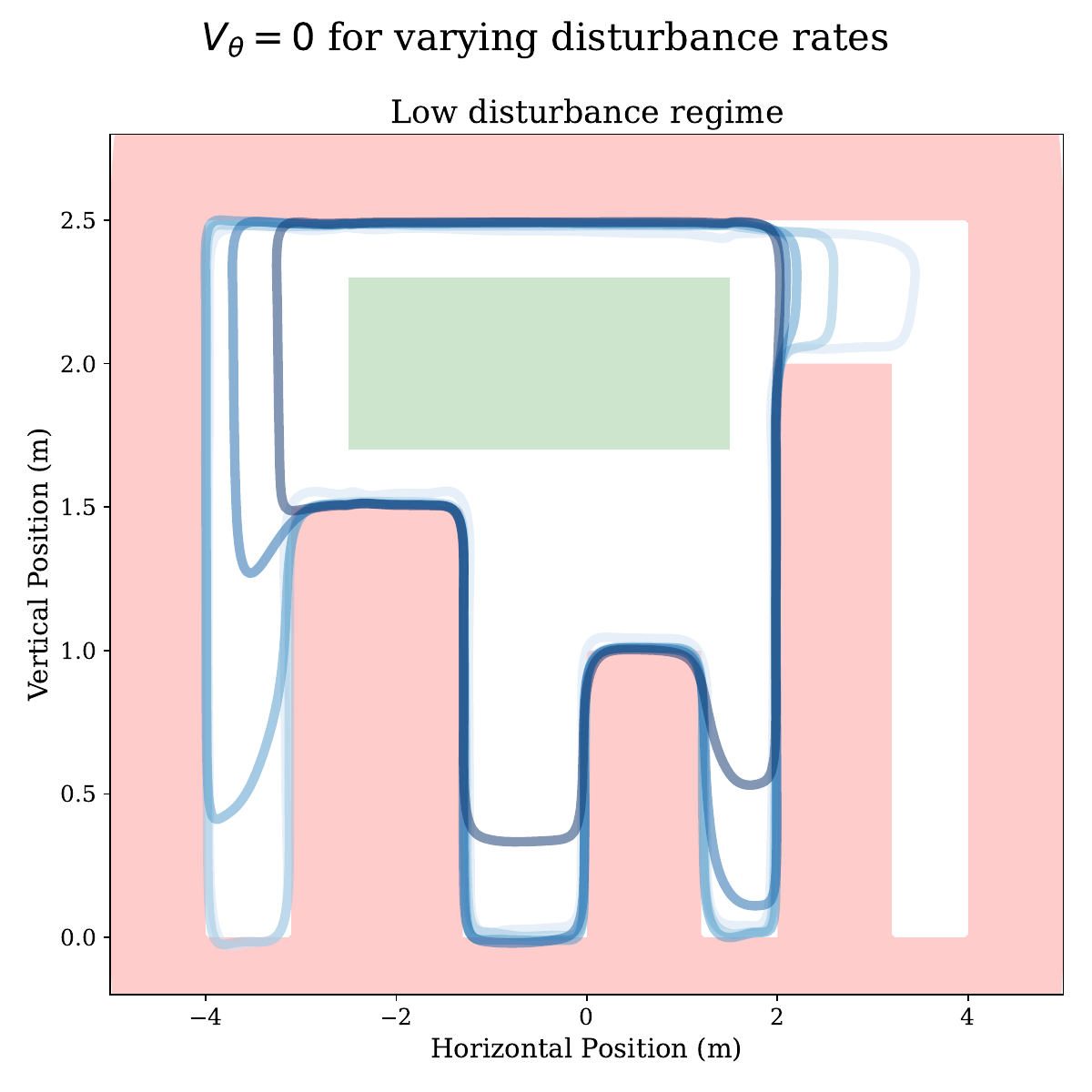}}
\end{subfigure}%
\begin{subfigure}[t]{0.33\linewidth}
  \centering
  \resizebox{\linewidth}{!}{\includegraphics{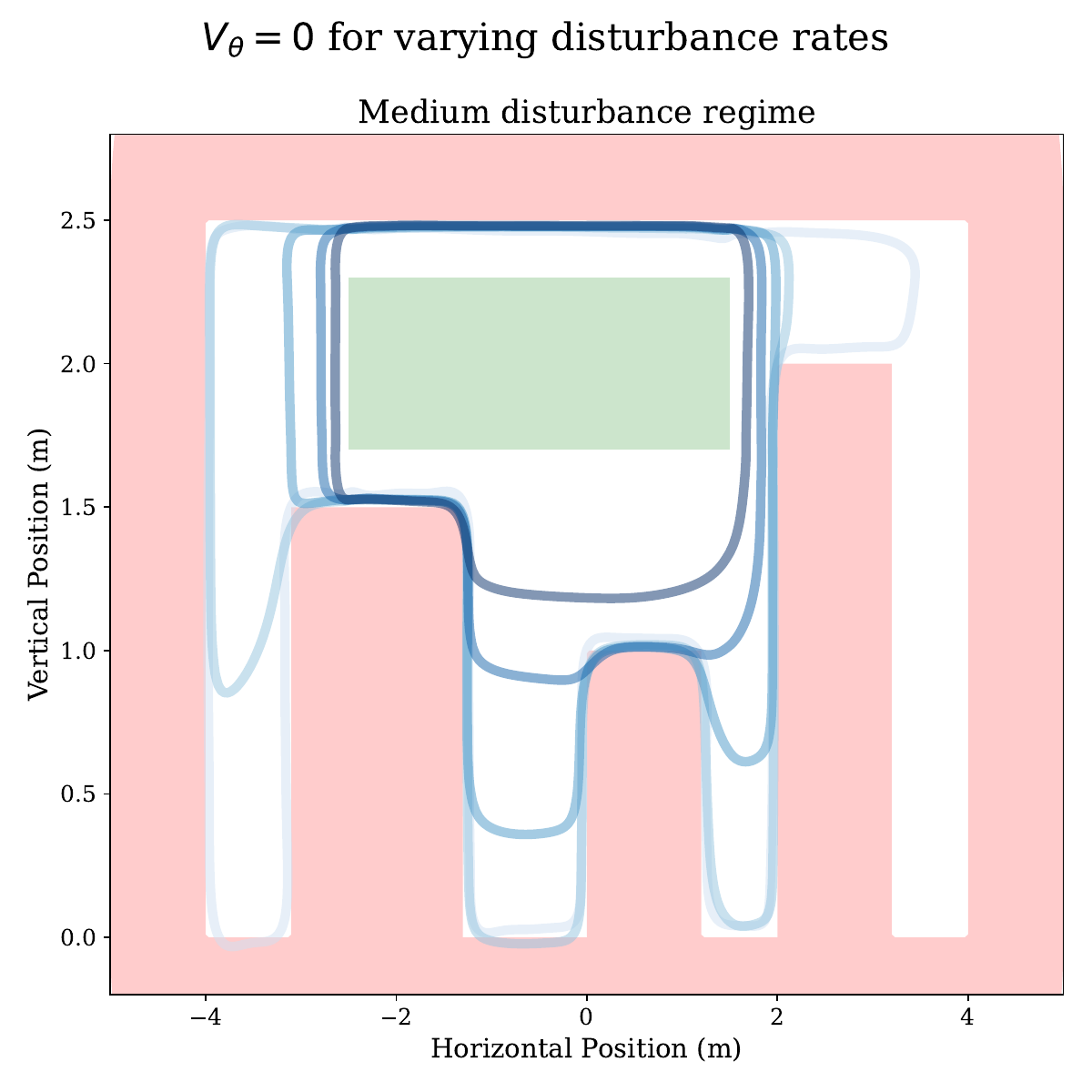}}
\end{subfigure}
\begin{subfigure}[t]{0.33\linewidth}
  \centering
  \resizebox{\linewidth}{!}{\includegraphics{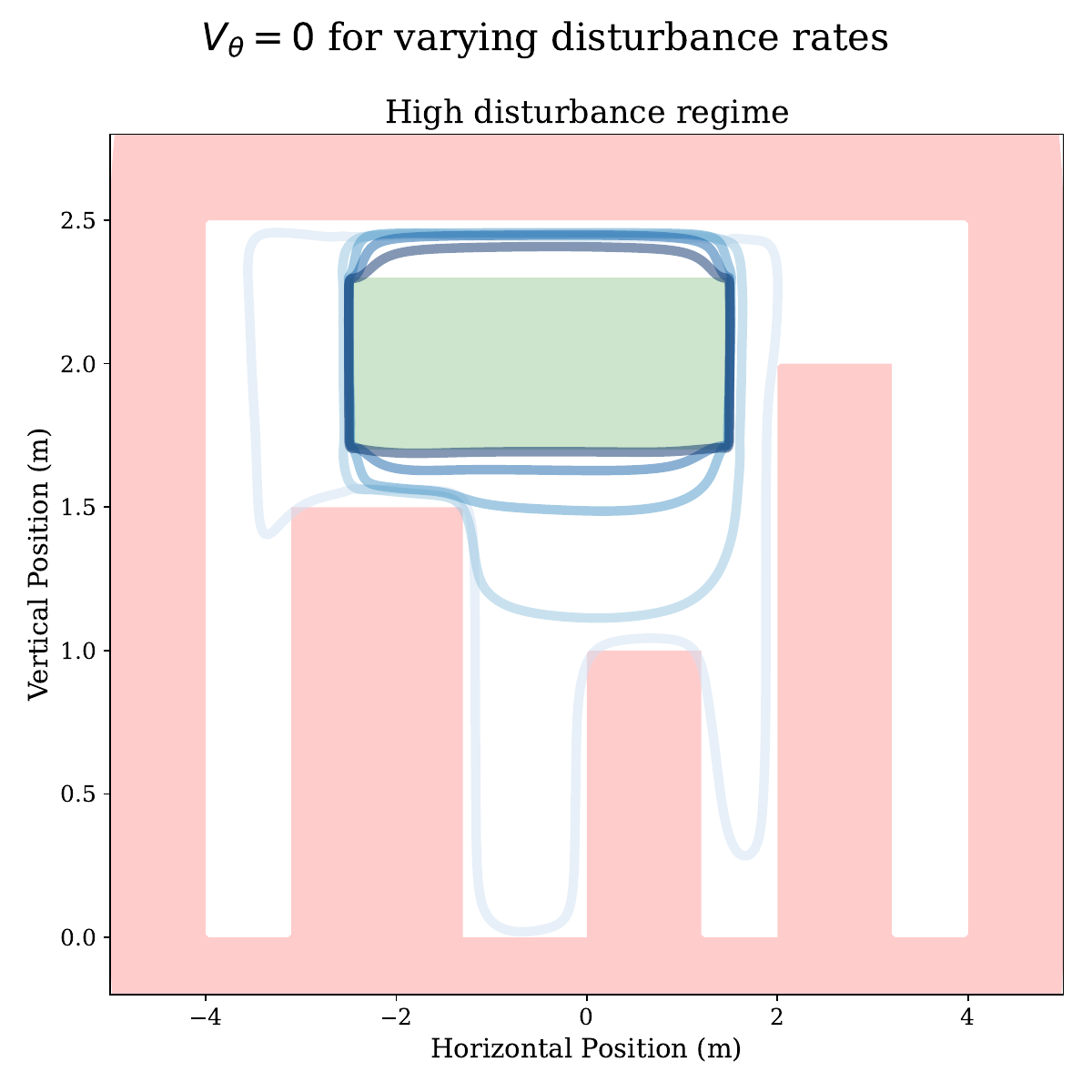}}
\end{subfigure}
\caption{The 0-level set of the learned value function for a fixed disturbance level $d$ for increasing (light to dark blue) disturbance rates $\dot{d}$. Left-to-right visualizes a low fixed disturbance level, a medium disturbance level, and a high disturbance level respectively. This is evaluated for the $v_x=0, v_z=0$ slice. $\dot{d}$ is encodes through the parameterized state, whereas $d$ is encoded through the time slice of the value function with $t=(d_\text{max} - d)/\dot{d}$. }
\label{fig:disturbancerates_levelsets}
\end{figure}

\textbf{Interpreting learned value functions}
Figures~\ref{fig:disturbance_levelsets} and~\ref{fig:disturbancerates_levelsets} visualize the 0-level sets of the learned value functions in the considered environment.
Specifically, the left, center, and right figures in Fig.~\ref{fig:disturbance_levelsets} showcase a varying disturbance magnitude level (from light blue to dark blue) for fixed low, medium and high disturbance rates. 
This showcases that even for high disturbance magnitudes as long as the disturbance rate is small (left) the safe region is relatively large, while for a high disturbance rates the safe region is much smaller (right).

Next, the left, center, and right figures in Fig.~\ref{fig:disturbancerates_levelsets} showcase a varying disturbance rate (from light blue to dark blue) for fixed low, medium, and high disturbance magnitudes. 
This figures showcases that as long as the current disturbance magnitude is small (left) the safe region is relatively large even for high disturbance rates, while for high current disturbance magnitudes (right) the safe region is only slightly larger than the control invariant set (right).

\section{Experimental Details}\label{sec:exp_design}

\subsection{Simulation Experiments} We consider a 4D drone dynamics model given by
\begin{align}\label{eq:4d}
\begin{bmatrix}
\dot{p}_x \\
\dot{p}_z \\
\dot{v}_x \\
\dot{v}_z
\end{bmatrix}
=
\begin{bmatrix}
v_x + d_1 \\
v_z + d_2 \\
g u_1 + d_3 \\
u_2 - g + d_4
\end{bmatrix},
\end{align}
where the control input is denoted by  $[u_1,\ u_2]^\top$ and the state vector is given by $[p_x,\ p_z,\ v_y,\ v_z]^\top$, with $p_y$ and $p_z$ denoting the position of drone, and $v_x$ and $v_z$ representing the corresponding velocity components\footnote{In the main body of the work we consider the state $[x, z, \dot{x}, \dot{z}]$ which overloads the notation on $x$therefore consider position $p$ and velocity $v$ with subscripts for its components}. The disturbances $d_1, d_2, d_3$ and $d_4$ represent wind, and $g$ is gravity.  We choose this low-dimensional model to highlight the performance of \algname{} for a ground truth value function using HJR (i.e. without the approximation errors inherent in learned approaches). Moreover, it allows us to demonstrate that our method generalizes across various strategies for computing the value function. While wind typically only affects the velocity components of the dynamics, we include the positional disturbances to circumvent the CBF (and value function) from learning to cancel out the disturbances directly in the control input (and thus effectively just reducing the control input bounds). Instead, by considering positional disturbances, the problem is interesting (while remaining tractable to solve with HJR methods).

\begin{figure}[ht]
 \centering
\includegraphics[width=9cm]{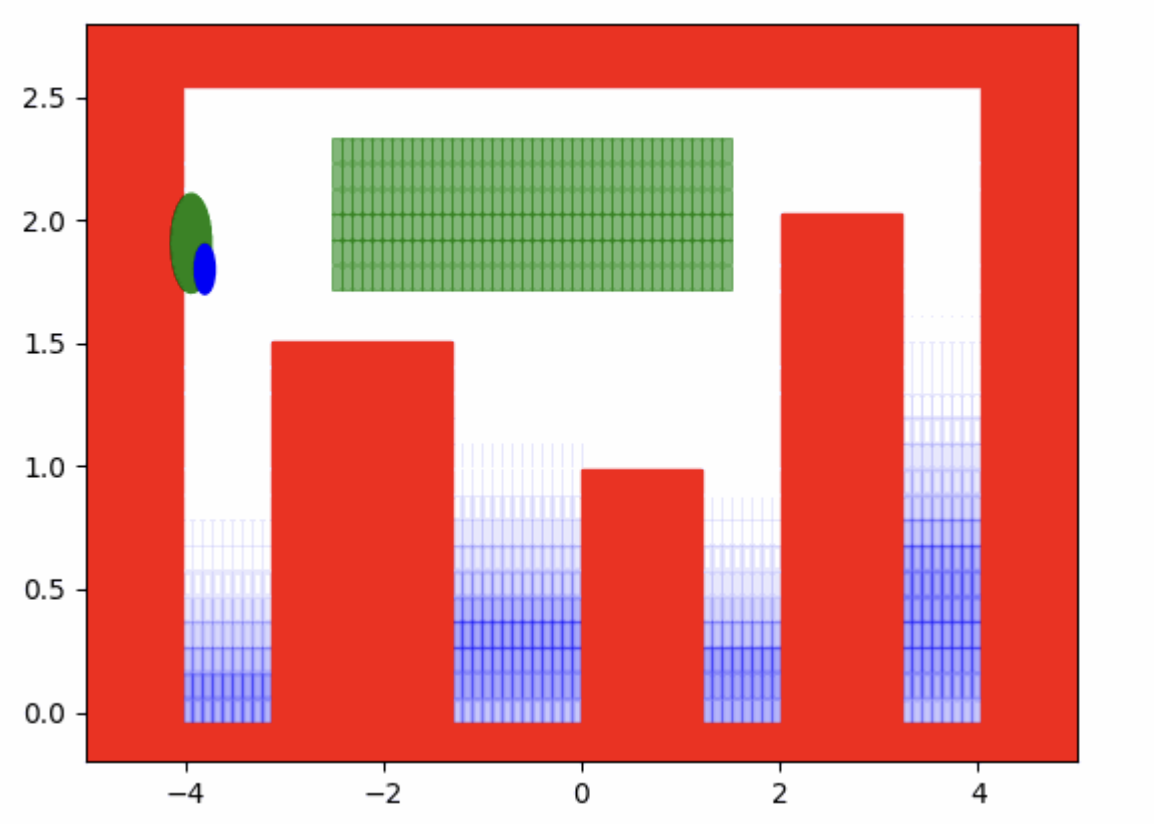}
\caption{Environment setup used in our simulations. This environment shows a quadcopter flying in a $2$D $p_x,p_z$ slice where $p_x$ is the horizontal axis and $p_z$ is the vertical axis. The quadcopter is denoted using a blue oval, while a current goal is shown using a green oval. All the obstacles in the environment, that denote unsafe regions for the quadcopter to enter, are shown in red. The safe control invariant target set that is used for our method is the translucent rectangle in free space shown in green. The wind between the urban canyon is shown in blue, where the darkness of the color corresponds to increased wind magnitude. For our experiments all wind disturbance is in the positive $x$ direction and negative $z$ direction. }
  \label{fig:env_setup}
\end{figure}

\textbf{Environment Setup.} We design a city-like environment with multiple tall buildings that the drone must avoid during its operation. The environment is defined in between $p_x \in [-5, 5]$ and $p_z\in[-0.2, 2.8]$, and includes three rectangular buildings located at: $(p_x, p_z) \in [-3.1, -1.3] \times [0, 1.5]$, $(p_x, p_z) \in [0.0, 1.2] \times [0.0, 1.0]$, and $(p_x, p_z) \in [2.0, 3.2] \times [0.0, 2.0]$. A spatial boundary restricts the drone to remain within $p_x \in [-4.0, 4.0]$ and $p_z \in [0.0, 2.5]$, and the velocity of the drone is constrained within $[-1.9, 1.9]$ for both $v_x$ and $v_y$. 
A safe target region, representing a designated flyover corridor above the cityscape, is defined as a rectangular set with $(p_x, p_z) \in [-2.5, 1.5] \times [1.7, 2.3]$.
To render the set approximately control invariant, both $v_x$ and $v_z$ are restricted to the range $[-1.0, 1.0]$ within the target region.
All obstacles and boundaries are indicated in red while the target region is depicted in green in  Figure~\ref{fig:env_setup}.

While our proposed method is capable of handling arbitrary disturbance models, we evaluate it in a specific urban canyon scenario. 
In this setup, wind is modeled as a deterministic disturbance with fixed direction in each state [1, -1, 1, -1]. 
The magnitude of the wind is bounded and increases toward the bottom of the canyons, capturing the channeling effects of urban geometry. 
The wind is shown in blue in Figure~\ref{fig:env_setup}. 
The opacity of the blue corresponds to the wind magnitude which increases as we descend (lower $p_z$). 
We model the increase in wind magnitude using an exponential function defined as
\begin{equation}
    f(W_{L}) = D \cdot \mathbf{W} \cdot 
    \left(1 - \exp\left(-\frac{r \cdot (W_L - W_{\text{min}})}{(W_{\text{max\_pos}} - W_{\text{min}})}\right)\right)
    \label{eq:exp_wind_fn}
\end{equation}
where $D$ denotes the disturbance factor. $ \mathbf{W}$ is the maximum wind value magnitude. \( r \) is the exponential ramp rate.  \( W_L \) is wind location which denotes the state of the robot where the wind is to be evaluated.\( W_{\text{min}} \) is the lower bound of the state range where the wind is defined. \( W_{\text{max\_pos}} \) is the position of the maximum wind value. This wind function defines the magnitude of the wind between $ W_{\text{min}} ,  W_{\text{max}} $ and the wind magnitude is $0$ outside these bounds. 
These wind functions are defined for every state dimension and composed. 

\textbf{Value Function Learning. } In simulation, we compare both the HJR and DR variants of our method against corresponding HJR and DR implementations of a baseline approach. 
Below, we outline the specific setup used for each method. 

\hspace{1cm}\textbf{DR Ours.}  
The DeepReach network is trained to approximate the reach-avoid value function over a time horizon from $0.0$ to $5.0$ seconds. The training time is approximately $2$ hours.

\hspace{1cm}\textbf{DR Naive.} As discussed in Section~\ref{sec:appendix_learning_ra}, due to challenges in obtaining a good performing value function for the avoid-only problem, we train DR to solve a reach-avoid problem, over a time horizon from $0.0$ to $5.0$ seconds. The training time is approximately $2$ hours.

\hspace{1cm}\textbf{HJR Ours} We compute the value function using dynamic programming via the JAX-based HJ Reachability Toolbox, \url{https://github.com/StanfordASL/hj_reachability}. 
The extended dynamics system, including disturbances, is of dimension $8$ and therefore cannot be computed directly using grid-based dynamic programming (limited to $4-5$ dimensions on a GPU, $6$ dimensions on a CPU, albeit taking hours to solve).
Therefore, to account for varying environmental conditions, we precompute $5$ value functions corresponding to disturbance rate magnitudes evenly spaced between zero and the maximum disturbance magnitude (with equivalent relative size for each disturbance dimension). 
During runtime, the safety filter associated with the smallest precomputed disturbance rate that exceeds the detected disturbance rate magnitude is selected and applied.

\hspace{1cm}\textbf{HJR Naive} This approach uses the same dynamic programming framework as \textbf{HJR Ours}. 
In this case, the five precomputed value functions correspond to different maximum disturbance magnitudes, evenly spaced between zero and the upper bound. 
At runtime, the safety filter associated with the smallest disturbance bound that exceeds the detected disturbance magnitude is selected for execution.

\hspace{1cm}\textbf{HJR Worst Case} This method computes a single safety value function using dynamic programming, based on the worst-case disturbance bounds observed in the environment.

\textbf{Experimental Results.}
We evaluate performance across 50 trajectories, each consisting of 10 random goals, with the drone navigating through the city environment under the wind, as summarized in Table \ref{tab:domain_comparison} in the main paper. 
The drone is controlled using a naive nominal controller based on Linear Quadratic Regulation (LQR), which does not account for obstacles in the environment. Goals are generated in a cyclical manner every 100 time steps to ensure diversity in target locations while remaining reachable by the nominal controller. Specifically, the first goal is placed above a randomly selected urban canyon, the second is set near the bottom of that same canyon, and the third is placed near the top. This three-step cycle is repeated to yield 10 goals per trajectory.

To test the robustness of our algorithm the wind field in the environment is randomized for every $10$ goal trajectory. 
For each wind field, the exponential rate governing the wind magnitude profile is sampled uniformly between $3$ and $7$, while the height at which the wind reaches its maximum is selected randomly between $0.1$ and one-third of the wind field height. 

\begin{figure*}[ht]
    \centering
    \begin{subcaptionblock}{0.5\textwidth}
        \includegraphics[width=\linewidth]{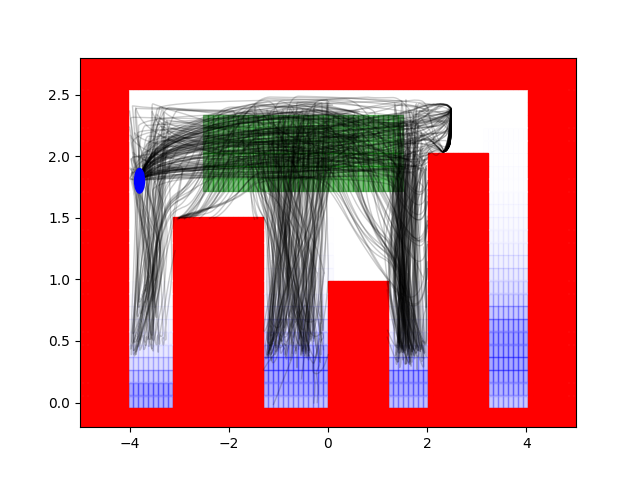}
        \caption{DR Ours}
        \label{fig:dr_ours_alltraj}
    \end{subcaptionblock}
    \hspace{-0.5em} % Reduced horizontal spacing
    \begin{subcaptionblock}{0.5\textwidth}
        \includegraphics[width=\linewidth]{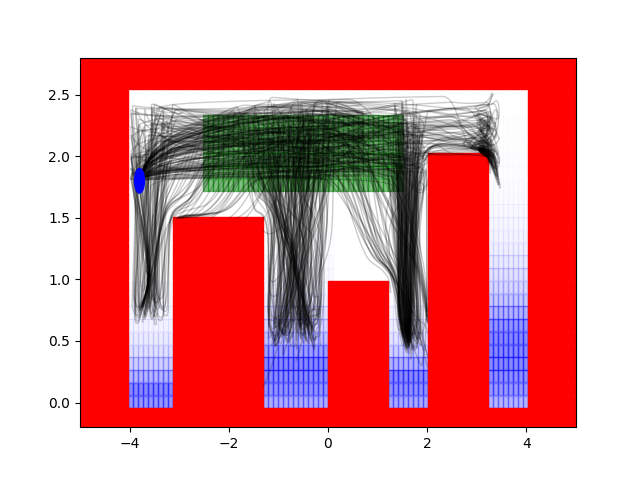}
        \caption{HJR Ours}
        \label{fig:hjr_ours_alltraj}
    \end{subcaptionblock}
    \caption{\textbf{Our Method Trajectories: }Figures showing $100$ of our trajectories moving between $15$ random goals through the city environment with randomized wind fields. 
    Notice that our safety filter prevents a majority of our trajectories from going too far down the urban canyon in response to increasing disturbance rate estimates. This leads to safer trajectories that fail less often. Fig. \ref{fig:dr_ours_alltraj} (on the left) shows these trajectories using our method with a learned value function using Deepreach with the 4 dimensional quadcopter model. 
    Fig.~\ref{fig:hjr_ours_alltraj} shows these trajectories using our method with value functions computed using HJR with the 4D quadcopter model.\ }
    \label{fig:ours_alltraj}
\end{figure*}

\begin{figure*}[ht]
    \centering
    \begin{subcaptionblock}{0.33\textwidth}
        \includegraphics[width=\linewidth]{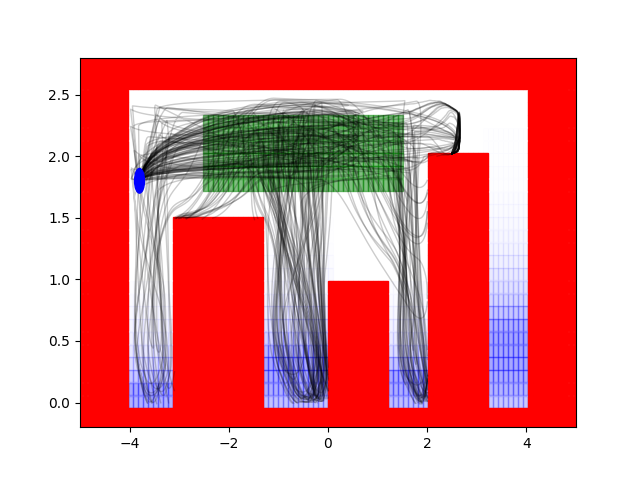}
        \caption{DR Naive}
        \label{fig:dr_naive_alltraj}
    \end{subcaptionblock}
    \hspace{-0.5em} % Reduced horizontal spacing
    \begin{subcaptionblock}{0.33\textwidth}
        \includegraphics[width=\linewidth]{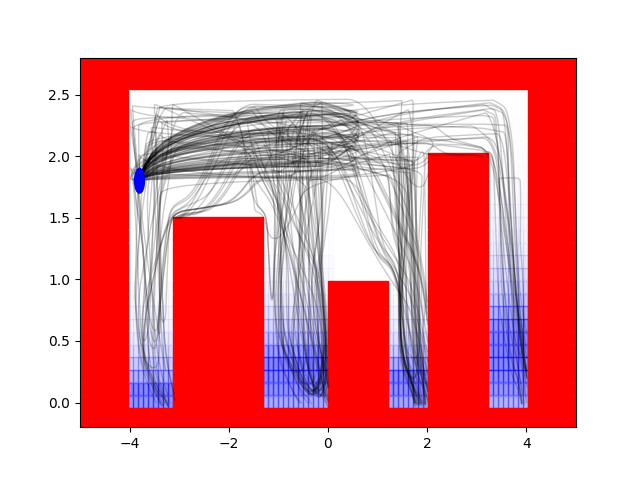}
        \caption{HJR Naive}
        \label{fig:hjr_naive_alltraj}
    \end{subcaptionblock}
    \hspace{-0.5em} % Reduced horizontal spacing
    \begin{subcaptionblock}{0.33\textwidth}
        \includegraphics[width=\linewidth]{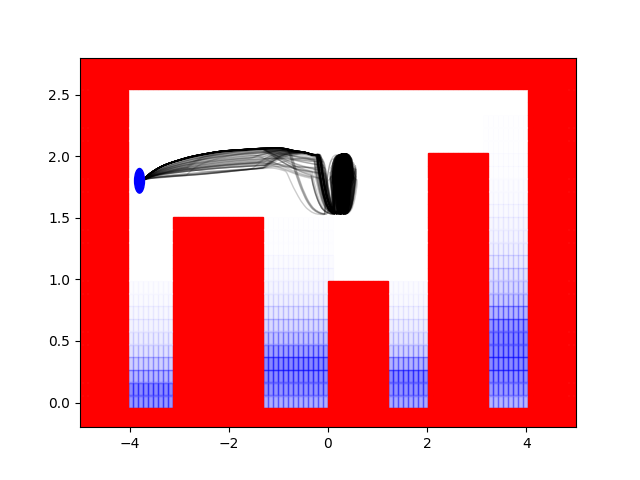}
        \caption{HJR Worst Case}
        \label{fig:hjr_worstcase_alltraj}
    \end{subcaptionblock}
    \caption{\textbf{Baseline Trajectories: }Figures showing $100$ of our trajectories moving between $15$ random goals through the city environment with randomized wind fields. Fig. \ref{fig:dr_naive_alltraj} (on the left) shows trajectories from the naive parameterized baseline using a value function learned with deepreach.  Fig. \ref{fig:hjr_naive_alltraj} (in the center) shows trajectories from the naive parameterized baseline using a value function computed using HJR. Notice that these trajectories fail to properly consider the unknown spatially varying disturbance and as a result go much further down the canyon and end up crashing. Notice that compared to our method showin in Figures \ref{fig:dr_ours_alltraj} and \ref{fig:hjr_ours_alltraj} the trajectory tracks are less dense. This is as a majority of the naive trajectories are cut short by the drones crashing whereas our method is able to continue for much longer. 
    Finally Fig. \ref{fig:hjr_worstcase_alltraj} shows the trajectories when using the HJR computed safety value function for the worst case disturbance bounds. 
    Only considering the worst case results in extremely conservative, non-performant trajectories that fail to move beyond a small set away from the obstacles.}
    \label{fig:baseline_alltraj}
\end{figure*}

We visualize all of the trajectories pertaining to our method in Figure~\ref{fig:ours_alltraj} and all the baseline trajectories in Figure~\ref{fig:baseline_alltraj}. 
From these figures we see the major impact of our method, where by properly reasoning about the potential increase in spatial disturbances our method does not descend far in the urban canyons and thus remains safe. 
Meanwhile, the naive baselines fail to reason about the spatially changing disturbances and ends up crashing (therefore cutting their trajectories short). 
This is also visible by looking at the density of the trajectory traces, where our ability to fly safely for longer is visualized through denser trajectory traces compared to the baselines that fail. 
The HJR Worst Case baseline, while safe, ends up being too conservative and is confined to a very small region of the environment. 

\textbf{Ablation Study} 
We conduct an ablation study on our recency based approach by varying $H = (1, 2, 5, 10)$, see Appendix Section~\ref{sec:algo_design} for more details on how its employed.  
We observe that changing the value of $H$ does not have a large affect on performance. 
In the case of HJR Ours, larger $H$ values reduce safety violations without adversarially affecting performance. 
We observe a slight increase in the crash rate for DR Ours, which we hypothesize is due to inaccuracies in the learned value function. 
Namely, the increased conservativeness causes the drone to spend more time hovering higher in the environment, which coincidentally corresponds to areas where our particular trained value function has a larger error compared to the HJR value function (ground truth). 
For all other methods the impact of $H$ is minimal.

\begin{figure*}[ht]
    \centering
    \includegraphics[width=\linewidth]{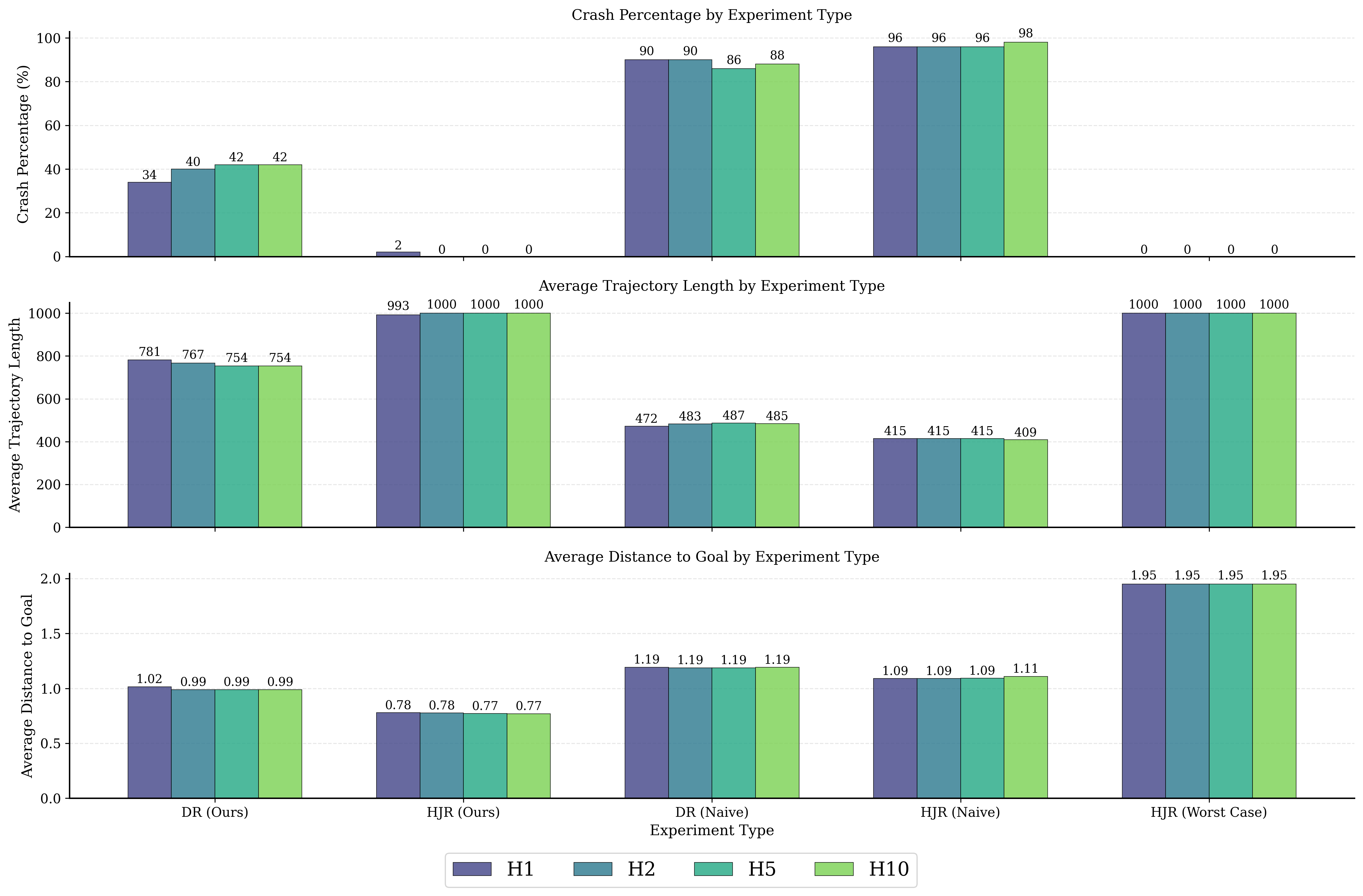}
    \caption{\textbf{H ablation experiment }This figure shows the result of the ablation experiment where $H$ is varied between $(1, 2, 5, 10)$. 
    Changing the value of $H$ does not have a large affect on performance. 
    In the case of HJR ours larger $H$ values reduce safety violations without adversarially affecting performance. 
    The increase in crash rate seen in DR Ours are attributed to inaccuracies in the learned value function. 
    For all other methods the impact of $H$ is minimal. }
    \label{fig:ablation_results}
\end{figure*}

\subsection{Hardware experiments} 
This hardware experiment setup discussion supplements the discussion in the main paper, Sec.~\ref{sec:result}
The hardware experiments are conducted in a planar $x-z$ plane, with the following state range $p_x \in [-4.0,4.0]$ and $p_z\in[0.0, 2.0]$. 
On a quadcopter, unlike in simulation, we cannot set the pitch rate $\theta_x$ directly.
Instead, we consider a cascaded model, inspired by~\cite{Herbert2021ScalableLO}, with as inputs the desired pitch rate $S_x$ and the combined thrust $T$.
The control inputs for the full system are $S_x$, $S_y$, $\dot{\psi}$, and $T$, with $S_y$ the desired roll rate and $\dot{\psi}$ the yaw rate.
The nominal controller is an LQR-based controller for $x_\text{LQR}=[p_x, v_x, p_y, v_y, p_z, v_z, \psi]$. 
Then, $S_{x,\text{nom}} = K_{p_x} (p_x - p_{x,\text{goal}}) + K_{v_x} v_x$, $S_{y,\text{nom}} = K_{p_y} (p_y - p_{y,\text{goal}}) + K_{v_y} v_y$, $\dot{\psi}_\text{nom} = K_\psi (\psi-0)$, and $T_\text{nom}=K_{p_z} (p_z - p_{z,\text{goal}}) + K_{v_z} v_z$.
Specifically, we set $K_{p_x}=K_{p_y}=K_{v_x}=K_{v_y}=-0.2$, $K_\psi=-20.0$ and $K_{p_z}=K_{v_z}=-10.0$
As our experiments are planar the nominal model considered for our CBF is $u_\text{nom} = [S_{x,\text{nom}}, T]$. 
The nominal desired roll attitude and yaw rate are directly fed into the system, with the objective of keeping $p_y=p_{y,\text{goal}}=0.0$ throughout the trajectory and keeping $\psi=0$.
We consider a 6D drone dynamics model given by
\begin{equation}\label{eq:6d}
    \begin{bmatrix}
\dot{p}_x \\
\dot{v}_x \\
\dot{\theta}_x \\
\dot{\omega}_x \\
\dot{p}_z \\
\dot{v}_z
\end{bmatrix}
=
\begin{bmatrix}
v_x \\
g \tan(\theta_x) + d_x - c_x v_x \\
-d_1 \theta_x + \omega_x \\
-d_0 \theta_x + n_0 S_x \\
v_z \\
k_T T_z - g + d_z
\end{bmatrix}
\end{equation}
where the desired pitch is given by $S_x$ and the state vector is given by $[p_x,\ v_x,\, \theta_x, \, \omega_x, \, p_z,\ v_z]^\top$, with $p_x$ and $p_z$ denoting the position of drone, $v_x$ and $v_z$ representing the corresponding velocity components, $\theta_x$ denothing pitch, and $\omega_x$ denoting pitch rate. 
The disturbances $d_x$ and $d_z$ represent wind, and $g$ is gravity.
Extending upon~\cite{Herbert2021ScalableLO}, we consider a drag term which we found to be necessary to achieve a good model fit.
The parameters $c_x, d_1, d_0, n_0, k_T$ that we used were fit using system-identification on a $1$ minute trajectory with random setpoints in $[p_x, p_y, p_z]$ updated every $6$ seconds, and are $c_x = 0.3$, $d_1 = 4.5$, $d_0 = 20.0$, $n_0 = 18.0$, and $k_T = 0.83$. 

\textbf{Environment Setup.} Our hardware experiments are conducted in an OptiTrack motion capture arena for precise state estimation. 
To emulate the simulation conditions, we construct a mock urban environment consisting of three tower-like obstacles built from stacked boxes. However, due to the space constraints of the flight arena the boundary's and each obstacle's $x$ positions are scaled by a factor of $0.8$ and $z$ positions are scaled by a factor of $0.75$. 
For the results shown in Figure~\ref{fig:realworld}, we introduce artificial disturbances into the state measurements to mirror the simulated dynamics. 
Specifically, we use the OptiTrack system to measure the drone's state in real time, evaluate the corresponding disturbance (using the same wind function employed in our simulation experiments), and add the resulting values into the velocity components along the $x$ and $z$ axes, $v_x$ and $v_z$.
This setup effectively spoofs the model to think it is perturbed by actual wind, thus allowing for consistent (non-turbulent) airflow, while mimicking the wind profile of urban canyons.
This setup enables consistent comparison between our proposed method and baselines under equivalent disturbance profiles.
The fans are placed in the environment for conceptual visualization only.
The reasons for not using fans are two-fold: 1) The airflow profile of radial fans causes a very rapid increase in wind disturbance at the edge of the fans, destabilizing the drone. 2) Measuring the wind disturbance without an airflow sensor is difficult and relies on single-step error measurements from the Optitrack system has too much variance to provide useful estimates. The drone platform itself has very limited compute and is purely used for state estimation using its IMU (in combination with the external Optitrack system) and sending input commands. A workstation (NVIDIA 4090 GPU, 64 GB RAM) runs the safety filter on a workstation and broadcasts the low-level (desired pitch and combined thrust) control commands. In addition a nominal controller ensures the yaw and roll of the quadcopter remain near $0$. 

\textbf{Value Function Training.} We evaluate different strategies for learning value functions under varying environmental conditions and disturbance models:

\hspace{1cm}\textbf{DR Ours.} We train a reach-avoid value function for the 6D quadrotor model in the urban environment, parameterized by the disturbance rate affecting the $x$ and $z$ velocity components, $v_x$ and $v_z$. The environment includes time-varying dynamics, with maximum disturbance magnitudes of $0.75$ in both velocity directions, and maximum disturbance rate of $1.5$ in both velocity directions. The time horizon is from $0.0$ to $5.0$ seconds and the value function is learned using the reach-avoid control-invariant loss~\eqref{eq:loss_and_hamiltonian_raci}.

\hspace{1cm}\textbf{DR Naive.} As in the simulation experiments, directly learning an avoid-only safety value function leads to poor convergence and suboptimal performance. 
    Instead, we train a reach-avoid value function under time-invariant dynamics with maximum disturbance magnitudes of $0.75$ in both velocity directions. The time horizon is from $0.0$ to $5.0$ seconds and the value function is learned using the reach-avoid control-invariant loss~\eqref{eq:loss_and_hamiltonian_raci}.

\textbf{Experimental results}
We evaluate DR Ours and DR Naive over 5 trajectories. We measure a trajectory as successful if there is no collision, see Table~\ref{tab:hw} for details and Fig.~\ref{fig:realworld} in the main paper for the visualization of the observed state trajectories in hardware for the trajectories.

\begin{table}[t]
\caption{Comparison of DR Ours approach against DR Naive in hardware (over 5 trajectories). We define a trajectory as a success if no crash occurs.}
\vspace{0.1cm}
\label{tab:hw}
\centering
\begin{tabular}{lcc}
\toprule
 & DR Naive & DR Ours \\
\midrule
Success Rate & 20\% & 100\% \\
\bottomrule
\end{tabular}
\end{table}
\section{\algname{}: Implementation details} \label{sec:algo_design}
The \algname{} algorithm (Alg.~\ref{alg:cap}) is extensively detailed in the main paper and algorithm block (Alg.~\ref{alg:cap}). 
However, the implementation of Line 2 of Alg.~\ref{alg:cap} (determining the disturbance magnitude and the disturbance magnitude rate) is nuanced, as explained below.

We consider a typical scenario where the control frequency is higher than the update rate $\Delta_d$ of the disturbance estimator. 
For each update to the current disturbance magnitude $\lvert d \rvert$, we estimate the directional derivative as $D_{\tilde{f}}d=\max\left\{0,\frac{\lvert d \rvert - \lvert d_\text{prev}\rvert}{\Delta_d}\right\}$, where $\lvert d_\text{prev}\rvert$ is the previous disturbance magnitude estimate. 
We consider only the disturbance magnitude as we consider the setting in which the maximum disturbance is bounded by a 0-centered hyperrectangle, i.e. $\mathcal{D}_{\max}=\{d\in\mathbb{R}^q \mid \lvert d \rvert \leq d_{\max}\}$. 

Dealing with decreasing disturbances requires extra care. 
Our formulation does not capture decreasing rates, instead conservatively assuming the disturbance magnitude stays constant. 
However, the value function $V(z,t)$ for state $z=[x,\dot{d}=0]$ only capture the constant but maximum disturbance magnitude $d_\text{max}$. 
Thus we bound $D_{\tilde{f}} d=\max\{D_{\tilde{f}} d, \frac{d_\text{max}}{t_\text{max}}\}$, where $t_\text{max}$ is the horizon for which the value function is learned or computed. 

Next, we discuss the implementation of the horizon $H$, which determines $\overline{d}$ and $\overline{D_{\tilde{f}}d}$. 
Consider the last disturbance measured is at time $t - \Delta_d \leq t_d\leq t$, with $t$ the current time, and we keep track of the last $H$ estimated disturbance rates. 
Then, we select the largest directional flow magnitude $\overline{D_{\tilde{f}}d} = \max_{h \in {0,..,H-1}}{D_{\tilde{f}}d(t_d-\Delta_d h)}$ along with the most recently estimated disturbance, $\overline{d}=\lvert d(t_d)\rvert $. 
For the naive baseline approaches, we keep track of the last $H$ estimated disturbances and use the disturbance with the max magnitude, i.e. $\overline{d}= \max_{h \in {0,..,H-1}}\lvert d(t_d-h)\rvert$. 
Lastly we clip $\bar{d}$ to $[0, d_\text{max}]$ and $\overline{D_{\tilde{f}}d}$ to $[0,\dot{d}_\text{max}]$ respectively. 

Lines 3-5 of Alg.~\ref{alg:cap} are explicit. 
They involve NN inference with gradients (for deepreach) or interpolating the value function and its gradients over a grid (for HJR) and solving a quadratic program (QP).
The deepreach implementation can be run at up to $500$Hz, but the HJR formulation is limited to approx. $40$Hz for our workstation (NVIDIA 4090 GPU, 64GB RAM). 

\end{document}